\documentclass{article}
\PassOptionsToPackage{numbers, compress}{natbib}
\usepackage[final,sglblindworkshop]{neurips_2025}

\usepackage[normalem]{ulem}
\usepackage{algpseudocode}
\usepackage{algorithm}

\usepackage[utf8]{inputenc} % allow utf-8 input
\usepackage[T1]{fontenc}    % use 8-bit T1 fonts
\usepackage{hyperref}       % hyperlinks
\usepackage{url}            % simple URL typesetting
\usepackage{booktabs}       % professional-quality tables
\usepackage{amsfonts}       % blackboard math symbols
\usepackage{nicefrac}       % compact symbols for 1/2, etc.
\usepackage{microtype}      % microtypography
\usepackage{xcolor}         % colors
\usepackage{graphicx}       % figures (add this?)
\usepackage{tikz}
\usepackage[many]{tcolorbox}
\usepackage{textcomp}
\usepackage{multirow}
\usepackage{wrapfig}
\usepackage{caption}
\usepackage{soul}
\usepackage{adjustbox}
\usepackage{enumitem}
\usepackage{caption}
\usepackage{tabularx}

\definecolor{CB_gray}{gray}{0.5}

\tcbset{prompt/.style={
    enhanced,
    size=fbox,
    boxrule=3pt,
    arc=2mm,
    auto outer arc,
    left=10pt,
    right=10pt,
    top=10pt,
    bottom=10pt,
    fontupper=\fontfamily{cmtt}\selectfont, 
    colback=CB_gray!10,
    colframe=CB_gray!25,
    coltitle=CB_gray!100, 
}}

\tcbset{vignette/.style={
    enhanced,
    size=fbox,
    boxrule=3pt,
    arc=2mm,
    auto outer arc,
    left=10pt,
    right=10pt,
    top=10pt,
    bottom=10pt,
    fontupper=\fontfamily{Avenir}\selectfont, 
    colback=CB_gray!10,
    colframe=CB_gray!25,
    coltitle=CB_gray!100, 
}}

\definecolor{codegreen}{rgb}{0,0.6,0}
\definecolor{codegray}{rgb}{0.5,0.5,0.5}
\definecolor{codepurple}{rgb}{0.58,0,0.82}
\definecolor{backcolour}{rgb}{0.95,0.95,0.92}

\usepackage{listings}
\usepackage{makecell}

\lstdefinestyle{mystyle}{
  commentstyle=\color{codegreen},
  keywordstyle=\color{magenta},
  numberstyle=\tiny\color{codegray},
  basicstyle=\ttfamily\footnotesize,
  stringstyle=\color{codepurple},
  breakatwhitespace=false,         
  breaklines=true,                 
  captionpos=b,                    
  keepspaces=true,                 
  numbers=left,                    
  numbersep=5pt,                  
  showspaces=false,                
  showstringspaces=false,
  showtabs=false,                  
  tabsize=2
}
\newcommand{\figurespace}{\vspace{-7mm}}

\def\Snospace~{\S{}}

\newcommand{\eg}{\textit{e.g.,}~}
\newcommand{\ie}{\textit{i.e.,}~}

\newcommand{\bg}{\texttt{BoxingGym}~}

\definecolor{coco1}{HTML}{D9E4EC}
\definecolor{coco2}{HTML}{B7CFDC}
\definecolor{coco3}{HTML}{6AABD2}
\definecolor{coco4}{HTML}{385E72}
\hypersetup{
    colorlinks=true,
    linkcolor=coco3,
    filecolor=coco4,      
    urlcolor=coco3,
    citecolor=coco3,
}

\usepackage[frozencache,cachedir=minted-cache]{minted}

\usepackage{csquotes}

\workshoptitle{Scaling Environments for Agents (SEA)}

\title{BoxingGym: Benchmarking Progress in Automated Experimental Design and Model Discovery}

\author{%
Kanishk Gandhi\thanks{Equal Contribution. Corresponding author: \texttt{kanishk.gandhi@stanford.edu}} \quad Michael Y. Li \footnotemark[1] \quad Lyle Goodyear \quad Agam Bhatia \AND
Louise Li \quad Aditi Bhaskar \quad Mohammed Zaman \quad Noah D. Goodman\\ \\
Stanford University
}

\begin{document}

\maketitle

\begin{abstract}
\vspace{-2mm}
Understanding the world and explaining it with scientific theories is a central aspiration of artificial intelligence research.
Proposing theories, designing experiments to test them, and then revising them based on data are key to scientific discovery.
Despite the promise of LLM-based scientific agents, no benchmarks systematically test their ability to propose scientific models, collect experimental data, and revise them in light of new data.
We introduce \bg{}, a benchmark with 10 environments for evaluating experimental design (\eg collecting data to test a scientific theory) and model discovery (\eg proposing and revising scientific theories).
To enable quantitative and principled evaluation, 
we implement each environment as a generative probabilistic model with which a scientific agent can run interactive experiments. 
These probabilistic models are drawn from various real-world scientific domains ranging from psychology to ecology.
To evaluate a scientific agent's ability to collect informative experimental data, we compute the expected information gain (EIG), an information-theoretic quantity which measures how much an experiment reduces uncertainty about the parameters of a generative model.
A good scientific theory is a concise and predictive explanation.
To quantitatively evaluate model discovery, we ask a scientific agent to explain their model and evaluate whether this explanation helps another scientific agent make more accurate predictions.
We evaluate several open and closed-source language models of varying sizes.
We find that larger models (32B) consistently outperform smaller variants (7B), and that closed-source models generally achieve better results than open-source alternatives.
However, all current approaches struggle with both experimental design and model discovery, highlighting these as promising directions for future research.
\footnote{Project: \href{https://github.com/kanishkg/boxing-gym}{https://github.com/kanishkg/boxing-gym}}

\end{abstract}

\vspace{-5mm}
\begin{flushright}
``To understand a system, you must perturb it.''\\
-- George Box \textit{(ad sensum)}
\end{flushright}
% TODO
% 
\vspace{-5mm}
\section{Introduction}
\vspace{-2mm}
\begin{figure}
    \centering
    \includegraphics[width=0.95\textwidth]{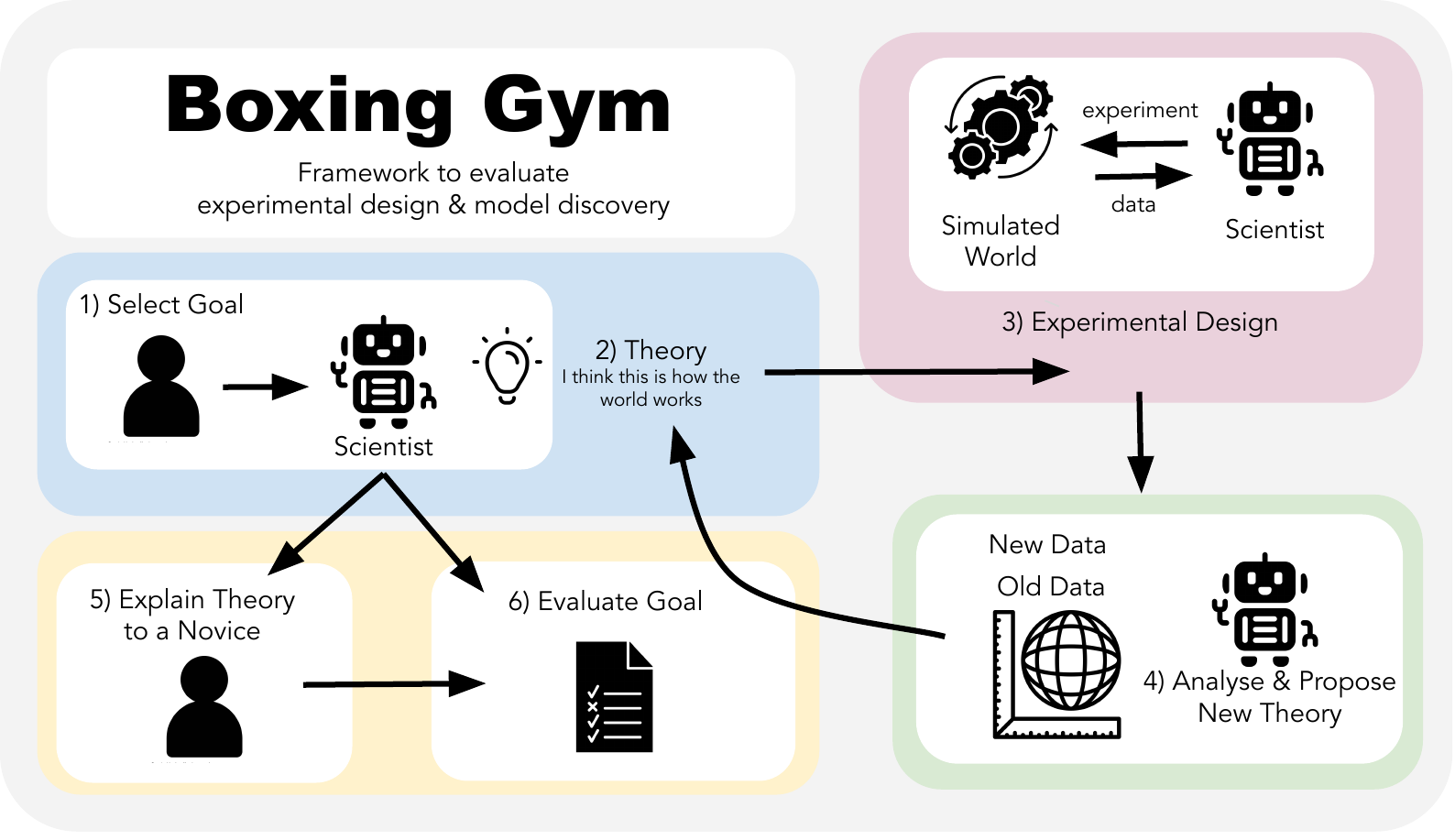}
    \caption{\textbf{Overview of \texttt{BoxingGym}.} The \bg Framework is designed to holistically evaluate experimental design and model discovery capabilities in the spirit of George Box \citep{box1976science}. 1) The process starts with a user defining a goal for the scientist agent. 2) The scientist formulates a theory. 3) This theory guides the experimental design, where the scientist interacts with a simulated world to gather new data. 4) The scientist then analyzes the new and old data to propose and refine theories. This iterative process continues for several iterations. 5) The scientist is then asked to explain the findings to a novice. 6) We evaluate the novice and the scientist by casting the goal as a prediction problem. }
    \label{fig:front}
\figurespace{}
\end{figure}

Helping humans understand the world (and themselves) by discovering scientific theories is a foundational goal of artificial intelligence research \citep{mccarthy1955proposal}.  Proposing theories about the world, conducting experiments to test them, and revising them based on data is central to this process \citep{box1976science}. Recent advances in large language models (LLMs), have shown promising potential for accelerating scientific discovery. LLMs have extensive scientific knowledge \citep{ai4science2023impact}, strong inductive reasoning capabilities \citep{wang2023hypothesis, qiu2024phenomenal}, and the ability to propose models of data \citep{li2024automated, li2024criticalcriticautomationlanguage, Castro2025.02.05.636732}. 
These promising results suggest that LLMs, functioning as autonomous agents, could be well-suited for experimental design (\ie collecting informative experiments to test scientific theories) and model discovery (\ie developing interpretable models based on experimental data).

Previous work has evaluated automated experimental design and model discovery in isolation \citep{foster19, foster21a, pmlr-v28-duvenaud13, li2024automated}. 
However, they are fundamentally coupled in real-world settings: scientists collect experimental data to build better models and better models inform better experiments.
While scientific agents are promising, there is currently no systematic way to evaluate an agent's ability to propose scientific models, collect experimental data, and revise them in light of new data. 
This motivates the need for a benchmark that evaluates an agent's capabilities holistically in an integrated scientific discovery pipeline.  

We outline the key desiderata for a framework that evaluates experimental design and model discovery: (1) The framework should enable the agent to \textit{actively experiment} with the environment without requiring the agent to perform time-consuming and resource-intensive real-world lab experiments. (2) Since scientific theories come in different forms, the framework should flexibly accommodate \textit{different representations of scientific theories}.
(3) The framework should evaluate experimental design and model discovery in an \textit{integrated} way. 
(4) Science is often \textit{goal-directed} or driven by an inquiry.
For example, a biologist might perform experiments with the goal of identifying cellular mechanisms underlying circadian rhythm in mammals. 
Our framework should allow users to specify high-level goals to guide the agent's discovery process. 
Our desiderata are inspired by the framework for scientific modeling introduced by George Box \citep{Box1962AUM,box80}, which emphasizes an iterative process of building models, designing experiments to test them, and revising them accordingly.

\begin{figure}[t]
\centering
\includegraphics[width=0.9\textwidth]{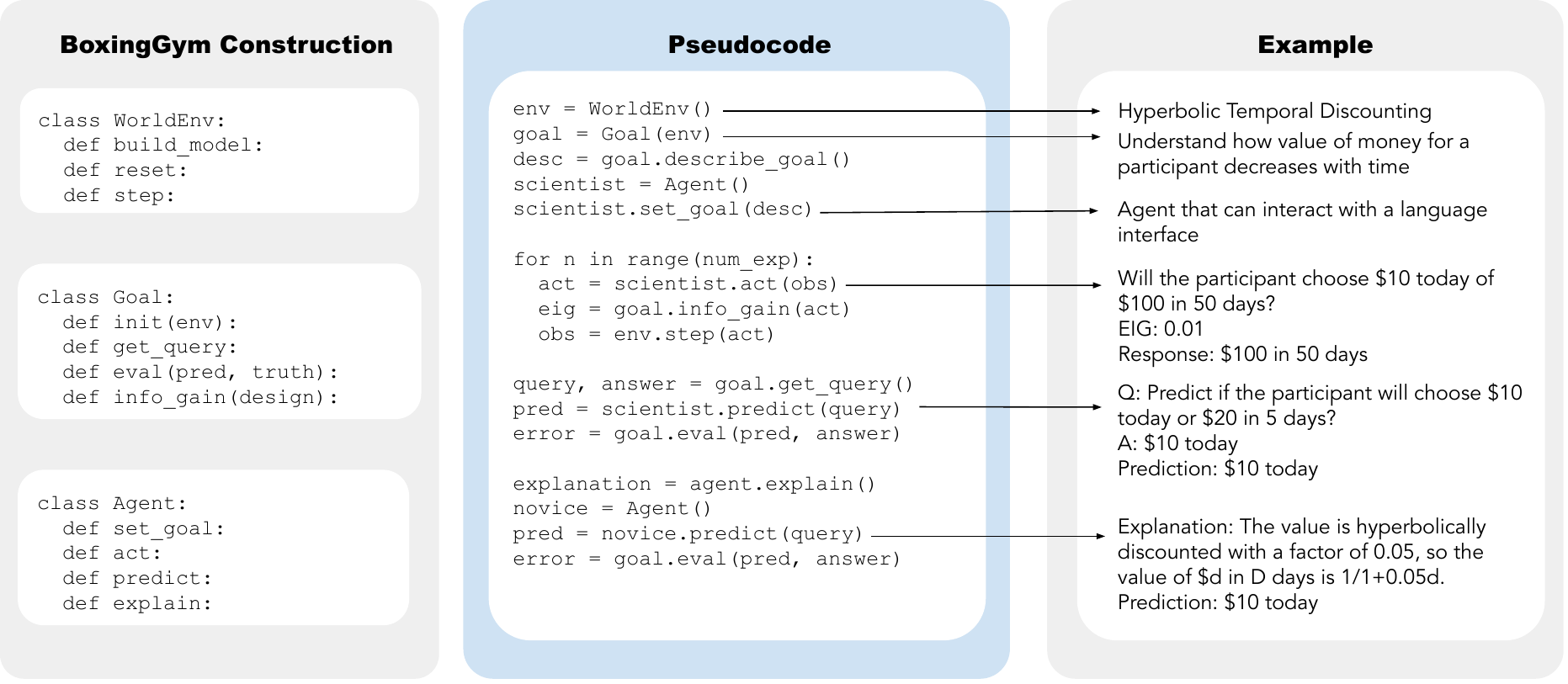}
\caption{\textbf{Python pseudocode examples.} \textbf{(left)} \bg is instantiated as modular classes and methods for the environment (WorldEnv), goals (Goal), and agents (Agent).
\textbf{(center)} Pseudocode illustrating the workflow of setting goals, performing experiments, predicting outcomes, and providing explanations. \textbf{(right)} An example, hyperbolic temporal discounting, where the agent predicts a participant's choice between immediate and delayed rewards and explains the concept to a novice.}
\label{fig:pseudo_code}
\figurespace{}
\vspace{1mm}
\end{figure}

To achieve these desiderata, we introduce \bg (\autoref{fig:front}) a flexible framework for evaluating experimental design and model discovery with autonomous agents. 
Our benchmark consists of 10 \textit{environments} grounded in real-world scientific models.
To enable agents to actively experiment, we implement each environment as a generative model. 
This key design choice makes simulating active experimentation tractable because it corresponds to sampling from the underlying generative model, conditioned on the experimental interventions. 
To accommodate various representations of scientific theories,  
all environments are designed with a flexible language based interface (\autoref{fig:pseudo_code}).
Finally, our environments can be instantiated with different goals, or intents for inquiry, that encourage the agent to adapt their experimentation towards accomplishing the goal (\eg understand the parameters underlying participant behavior in a psychology study) by specifying the goal in language.

We introduce principled evaluation metrics that measure the quality of experiments and discovered models. To evaluate experimental design, we draw from \textit{Bayesian optimal experimental} (BOED) design \citep{rainforth2017opportunities} and use \textit{expected information gain} (EIG) to measure the informativeness of an experiment. EIG captures how much an experiment reduces uncertainty in the parameters of a generative model and, importantly, this measure complements our decision to implement environments as generative models. To evaluate model discovery, we take inspiration from the fact that science is a communicative endeavor. We propose a \textit{communication-based} evaluation strategy: we ask a scientist agent to distill their experiments into a natural language explanation and evaluate how much that explanation empowers a novice agent, who does not have access to the experiments conducted by the scientist, to make accurate predictions about the environment. 

We evaluate several open and closed-source language models ranging from 7B to 32B parameters. 
We find that larger models consistently outperform smaller variants, and closed-source models generally achieve better results than open-source alternatives.
We also evaluate Box's Apprentice \citep{li2024automated}, which augments language models with statistical modeling capabilities, but find that this augmentation does not reliably improve performance. 
Notably, we observe substantial variation in difficulty across environments, which remaining challenging even for the strongest models. 
Promisingly, some environments show clear performance improvements with model scale.
These results highlight significant opportunities for improving automated scientific reasoning.
% We introduce principled evaluation metrics that measure the quality of experiments and discovered models. 
% To evaluate experimental design, we draw from \textit{Bayesian optimal experimental} (BOED) design \citep{rainforth2017opportunities} and use \textit{expected information gain} (EIG) to measure the informativeness of an experiment.
% EIG captures how much an experiment reduces uncertainty in the parameters of a generative model and, importantly, this measure complements our decision to implement environments as generative models.
% To evaluate model discovery, we take inspiration from the fact that science is a communicative endeavor.
% We propose a \textit{communication-based} evaluation strategy: we ask a scientist agent to distill their experiments into a natural language explanation and evaluate how much that explanation empowers a novice agent, who does not have access to the experiments conducted by the scientist, to make accurate predictions about the environment.
% Finally, as initial steps, we implement and evaluate two baseline agents: (1) a purely language-based agent (2) a language agent that builds statistical models of the data and uses the model to aid their predictions \citep{li2024automated, li2024criticalcriticautomationlanguage}.
% These tasks are challenging for both agents and highlight fertile ground for further research.
% \vspace{-1mm}

\vspace{-3mm}
\section{Related Works}
\vspace{-2mm}
\paragraph{Optimal Experimental Design.} 
Bayesian optimal experimental design (BOED) is a principled framework for designing maximally informative experiments across various disciplines \citep{shabobo2013, boed_review, myungh}.
While theoretically appealing, BOED's practical implementation is challenging due to the intractability of information gain metrics like expected information gain (EIG).
Although several methods \citep{rainforth2017opportunities, foster19, foster21a} exist to approximate EIG, they assume the data follows a fixed generative model—limiting their utility when model revision is needed as new data is collected.

\vspace{-3mm}
\paragraph{Automated Model Discovery.}
Automated model discovery from data has been a long-standing goal in AI, aiming to build interpretable models that capture underlying patterns in data—from physical laws \citep{Bongard2007AutomatedRE, mckinney_phys_review} to nonparametric regression \citep{pmlr-v28-duvenaud13}. Recent work  \citep{li2024automated, li2024criticalcriticautomationlanguage} has integrated language models into this process, leveraging their ability to both propose and critique candidate models, demonstrating their potential as tools for automated model discovery. This work highlights the potential of using language models as a powerful tool for model discovery.

% Automated model discovery from data has been a long-standing goal in AI, with the aim of building interpretable and explainable models that capture underlying patterns and relationships in data. Researchers have developed systems for discovering physical laws \citep{Bongard2007AutomatedRE, mckinney_phys_review}, performing nonparametric regression \citep{pmlr-v28-duvenaud13}, and conducting unsupervised learning \citep{Grosse2014ModelSI}.
% Recently, \citet{li2024automated, li2024criticalcriticautomationlanguage} proposed integrating language models into an automated statistical model discovery pipeline that involves both proposing candidate models and critiquing them, 
% leveraging their ability to generate symbolic expressions and critique their own reasoning. 

\vspace{-3mm}
\paragraph{Reasoning and Exploration with LLMs.}
Language models have shown promising capabilities in both deductive reasoning (deriving consequences from hypotheses) \citep{saparov2024testing, saparov2022language, poesia2023certified} and inductive reasoning (inferring hypotheses from observations) \citep{wang2023hypothesis, qiu2024phenomenal}. While reinforcement learning has improved LLMs' reasoning abilities \citep{jaech2024openai, guo2025deepseek, gandhi2025cognitive, havrilla2024teaching}, these advances have primarily focused on deterministic, verfiable systems rather than the stochastic data typical in scientific discovery.
Efficient exploration and information-seeking are crucial for experimental design and model building. Recent work \citep{nie2024evolve,min2021metaicl,gandhi2024stream,gandhi2023strategic,schultz2024mastering,lehnert2024beyond} has investigated in-context exploration strategies and shown how language models can learn how to search and explore directly through sequence modeling, developing effective search strategies in language.

\vspace{-3mm}
\paragraph{Interactive Environments.}
Drawing inspiration from established reinforcement learning principles \citep{brockman2016openai, mnih2013playing}, \bg adopts the modularity and simplicity of classic environments like OpenAI Gym while shifting focus to evaluation rather than agent training. While recent work has expanded interactive benchmarks to language agents ---spanning tasks from software debugging \citep{jimenez2024swebench} to automated scientific research\citep{nathani2025mlgym, lu2024ai}, our work advances this direction by introducing a principled framework for evaluating language agents' capabilities in iterative experimental design and model discovery.

% \citet{rainforth2017opportunities} propose a nested Monte Carlo estimator and \citet{foster19} estimate the EIG using a variational approximation. 
% Subsequent work has scaled BOED to sequential settings by learning an amortized design network \cite{foster21a}.
% While these methods are significant technical innovations, they typically assume that the data is well-approximated by an assumed generative model.
% However, in many practical settings, we also want to revise our model as we collect new data. 
\vspace{-3mm}
\section{Boxing Gym}
\vspace{-1mm}
\subsection{Problem Formulation.}
% % \vspace{-1mm}
We formalize experimental design and model discovery using probabilistic modeling and Bayesian optimal experimental design (BOED). 
In \bg, each environment is implemented as a generative model defining a joint distribution over the experimental outcome $y$, experimental design $d$, and unobserved parameters $\theta$.
This joint distribution is defined in terms of a prior distribution over $\theta$, $p(\theta)$ and a \textit{simulator} $p(y | \theta, d)$ which is a model of the experimental outcome $y$ given parameters $\theta$ and design $d$.
For example, in a psychology experiment, $\theta$ could be the parameters of a behavioral model of participants, $d$ could be the questions posed to participants, and $y$ could be the participant's response to $d$. 
Running an experiment corresponds to choosing a design $d$ and observing a sample $y$ from the marginal predictive distribution conditioned on that design, \ie $y \sim p(y | d) = E_{p(\theta)}[p(y | \theta, d)]$) \footnote{In the sequential setting, we replace the prior $p(\theta)$ with the posterior $p(\theta | y, d)$.}.

\subsection{Evaluation}\label{sec:metrics}
% \vspace{-1mm}
% We now discuss how we quantitatively evaluate a scientific agent's capabilities in 1) experimental design and 2) model discovery. 

\subsubsection{Evaluating experimental design via Expected Information Gain}\label{sec:eig}

To evaluate experimental design, we take inspiration from the Bayesian OED literature \citep{foster19,foster21a}. 
Crucially, our choice to implement environments as generative models enables us to leverage this literature.
For each domain, we have an underlying predictive model $p(y| \theta, d)$. 
% Here, $y$ corresponds to an experimental outcome, $\theta$ corresponds to parameters of the model, and $d$ corresponds to a particular experimental design.
% For example, consider hyperbolic temporal discounting a model from behavioral economics for modeling how people weigh immediate rewards against delayed ones 
% $y$ is participant's binary decision of whether to accept a delayed or immediate reward, $\theta$ consist of the parameters that dictate how a participant trades off between immediate and delayed rewards, and $d$ consists of a immediate reward value, delayed reward value, and the number of days that the reward is delayed by.
We quantify the \textit{informativeness} of a design $d$ through the expected information gain (EIG), that measures the reduction in posterior uncertainty about the model parameters $\theta$ after running an experiment $d$. 
Below, $H$ is the Shannon entropy.
\begin{align*}
\text{EIG}(d) = \mathbb{E}_{p(y|d)} \left[ H[p(\theta)] - H[p(\theta|y, d)] \right]
\end{align*}
Since the EIG is typically not available in closed-form, we use a Nested Monte Carlo estimator
\noindent
\makebox[\textwidth]{\parbox{1.3\textwidth}{%
\begin{align*}
\hat{\mu}_{\text{NMC}}(d) = \frac{1}{N} \sum_{n=1}^{N} \log \left( \frac{p(y_n|\theta_{n,0}, d)}{\frac{1}{M} \sum_{m=1}^{M} p(y_n|\theta_{n,m}, d)} \right) \quad \text{where} \quad \theta_{n,m} \overset{\text{i.i.d.}}{\sim} p(\theta), \; y_n \sim p(y|\theta = \theta_{n,0}, d)
\end{align*}
}}
We chose this estimator because it is a consistent estimator of the true EIG \citep{rainforth2017opportunities} and is straightforward to implement.
EIG measures the value of an experiment under the assumption that the true distribution of experimental outcomes is modeled by $p(y | d)$. 
In general, this assumption is not true, but EIG is still a useful measure since we generate data from an underlying model in our benchmarks.
% % \vspace{-2mm}
% \ndg{need to talk about the assumption here that the space of models is  given by the structure of the true model. this is clearly wrong (space of models is something like all models), so we should argue why it is still useful.}

% \ndg{i think you never talked about evaluation from goal prediction? subsection here?}

% \michael{prior/no prior}
\subsubsection{Evaluating model discovery via communication}\label{sec:comm}
\vspace{-2mm}
To evaluate the quality of a model, we use standard model evaluation metrics (\eg prediction MSE) and a communication-based metric that takes advantage of the natural language interface. 
In particular, a \textit{scientist agent} interacts with an environment through experiments. 
After these experiments, we ask the scientist agent to synthesize their findings through an \textit{explanation}.
We then evaluate how much that explanation enables a \textit{novice} agent to make more accurate predictions about the environment without any additional experiments. 
Since a good explanation is both \textit{predictive} and \textit{parsimonious}, we set a token limit on the explanation. Crucially, this evaluation method can accommodate different forms of scientific theories. 
In our experiments, we ask the scientist agent to produce a statistical model and then distill the model into a natural language explanation to guide the novice agent.

\subsubsection{Evaluating goals via prediction}
\vspace{-2mm}
To evaluate success at achieving a specific goal (\eg how do the populations of predator and prey change with time) we employ a prediction target (\eg predict the population of predators at a particular time) and calculate a standardized prediction error.  First, we compute the error between the predicted and true values. Then, we standardize this error with respect to the prior predictive mean, which is obtained by assuming a uniform prior over the design space. 
Specifically, for each domain, we sample a design $d$ uniformly from the design space and a parameter $\theta$ from the prior distribution $p(\theta)$. We then generate samples from the predictive model $p(y| \theta, d)$ and average over multiple $d$ and $\theta$ to obtain the prior predictive mean $\mu_{0}$ and variance $\sigma_0$. Let $\{y_i\}_{i=1}^n$ be the ground truth outputs for inputs $\{x_i\}_{i=1}^n$.
and let $\{\hat{y_i}\}_{i=1}^n$ be the predictions of the agent. The standardized prediction error is then calculated using these quantities, providing a measure of the agent's performance relative to the prior predictive mean. We use a domain-specific function $f$ computing the discrepancy between a prediction $\hat{y_i}$ and ground truth value $y_i$ (\eg MSE). 
We compute the errors $\epsilon_i = f(\hat{y_i}, y_i)$ and $\epsilon_{\mu_0} = f(\mu_0, y_i)$.
Finally, we compute the standardized error as
$\frac{\epsilon_i - \epsilon_{\mu_0}}{\sigma_{0}}$.
Crucially, since this metric is computed with respect to the prior predictive, this metric can be negative. 
\subsection{Design Decisions in Constructing \bg}
\vspace{-2mm}
We outline the key design decisions of \bg that allow it to capture key aspects of scientific discovery within a flexible, simulated, and extensible environment.
\vspace{-0.5mm}

\paragraph{Discovery via active experimentation.}  The agent actively interacts with the environment by conducting experiments, reflecting the real-world coupling of experimentation and model discovery. This approach assesses the agent's ability to gather relevant data and refine its models based on experimental results.
\vspace{-0.5mm}

\paragraph{Real-world scientific models.}
Our environments are grounded in real-world scientific models from several domains, ensuring the benchmark tests the agent's ability to handle realistic scenarios. We implement these environment as \texttt{pymc} generative models to make active experimentation an automatic and tractable process.

\vspace{-0.5mm}

\paragraph{Goal-driven discovery.}
Each environment has a specific goal, mirroring the inquiry-driven nature of scientific research. This encourages the agent to engage in targeted experimentation.
\vspace{-0.5mm}

\paragraph{Language-based interface for experiments.}
We use a language-based interface for our experiments because it's flexible (\ie scientific domains can generally be described in language), easily integrates with LLMs, and interpretable to humans.
\vspace{-0.5mm}

\paragraph{Emphasis on Measuring Discovery with Explanations.}
\bg places a strong emphasis on measuring the quality of the agent's discoveries through the explanations it can provide after experimentation (\autoref{sec:comm}). This design decision is motivated by two considerations. From a theoretical perspective, science is fundamentally about developing better theories, and scientific theories are explanations of observed phenomena. From a practical perspective, communicating findings to the broader scientific community is an essential aspect of scientific research.  
By using language, we do not have to commit to a particular representation of a scientific theory. 
We illustrate this flexibility, by showing how different representations can be easily integrated within our method for measuring natural language explanations.
\vspace{-0.5mm}

\paragraph{Extensible/modular environments for benchmarking agents.} \bg is easily extensible and modular, enabling researchers to integrate new environments and test different agents with minimal effort. 
We illustrate this in \autoref{fig:pseudo_code} which provides a pseudo-code example of how to implement a new environment and goal in \bg.

\subsection{Domains}

\bg consists of 10 environments (see \autoref{app:domains} for full details) that cover a range of scientific domains and test different aspects of experimental design and model discovery. Some environments are designed to test optimal experiment design, while others focus on model discovery or involve simulated neuro-symbolic human participants.

\paragraph{Location finding.} \citep{foster21a}  In an $n$-dimensional space with $k$ signal-emitting sources, the scientist measure signals at any grid location. Goals include predicting the signal at any point or locating the sources.

\paragraph{Hyperbolic temporal discounting.} \citep{foster21a}  The scientist observes a participant's choices for different immediate rewards ($ir$), delayed rewards ($dr$), and delay periods ($D$ days) \autoref{fig:pseudo_code} (right). Goals include predicting choices of a participant or discount factors.

\paragraph{Death process.} \citep{foster21a} A disease spreads at an infection rate. The scientist can measure the number of infected individuals at different points of time to predict future infections or the infection rate.

\paragraph{Item Response Theory (IRT).} \citep{rasch1993probabilistic} In this environment, there is a set of students and a set of questions. The experimenter can observe the correctness of a student's response to a particular question. The goal is to discover the underlying model that relates student ability and question difficulty to the probability of a correct response.
% \vspace{-0.5mm}

\paragraph{Animal growth curves.} \citep{Magnusson_posteriordb_a_set_2023}
An experimenter can observe the length of a dugong at a particular age. The goal is to discover the underlying growth model of dugongs.
% \vspace{-0.5mm}

\paragraph{Population growth dynamics.} \citep{Magnusson_posteriordb_a_set_2023} An experimenter can observe the population of peregrines at a particular point in time. The goal is to discover the underlying population dynamics model. This is tested by asking the experimenter to predict population dynamics at a particular point in time.
% \vspace{-0.5mm}

\paragraph{Mastectomy Survival analysis.} \citep{cox2018analysis} The experimenter can observe if a patient is alive after a mastectomy, including metastasis status and time since surgery. The goal is to predict survival probabilities for new patients.
% \vspace{-0.5mm}

\paragraph{Predator-Prey dynamics.} \citep{volterra1928variations} This simulates predator-prey populations over time. The goal is to discover models like the Lotka-Volterra equations to predict future populations.
% This environment simulates predator-prey dynamics. The experimenter can observe the populations of predators and prey at different time points. The goal is to discover the underlying model (lotka-volterra equations) governing the population dynamics. The goal is tested by asking the experimenter to predict the population of predator and prey at a specific time.

% \textbf{Environments with simulated neuro-symbolic human participants:}
% \vspace{-0.5mm}

\textbf{Emotion from outcome.} \citep{ong2015affective} Participants guess a player's emotions after a gambling game's outcome. The experimenter designs games with varied probabilities and prizes to model how participants judge the emotions of a player from outcomes. Human participants are simulated using a probabilistic model translated into natural language by a language model.

\textbf{Moral Machines.} \citep{awad2018moral} Participants face moral dilemmas, choosing which group an autonomous car should save. Experimenters manipulate group compositions and required actions to model moral decision-making. Human participants are simulated with a probabilistic model, and their actions are translated into natural language by a language model.

% This environment presents a moral dilemma similar to the trolley problem. An autonomous car with a set of people (group 1) is heading towards a barricade or a second group of people (group 2). If the car hits the barricade, group 1 will be harmed, and if it swerves, group 2 will be harmed. The human participant must choose which group to save. The groups can have different numbers of people and people from different demographics, and also pets, resulting in 39 million possible experiment configurations. The experimenter can choose the composition of the two groups and whether an active action is required. The human participant's actions are predicted by a simulated probabilistic model and translated into natural language using a language model. The experimenter's goal is to discover the underlying model that relates the group compositions and the requirement for active action to the human participant's judgement.

% These diverse environments enable the comprehensive evaluation of an agent's experimental design and model discovery capabilities across a wide range of scientific domains and problem types.
% \vspace{-3mm}
\vspace{-1mm}
\section{Experiments}
\vspace{-3mm}

We conduct experiments to evaluate the performance of two baseline agents on \bg. Our goal is to assess their ability to perform experimental design and theory building across a diverse set of environments. We benchmark two types of agents: a standard language model (GPT-4o, \citet{OpenAI_2024}) and a language model augmented with symbolic reasoning capabilities (Box's Apprentice). 

\paragraph{LLM Agent.} We consider 6 LLMs, GPT-4o \citep{OpenAI_2024}, Claude-3.7-sonnet \citep{anthropic2024claude}, Qwen-2.5-32b-instruct, Qwen-2.5-7b-instruct \citep{qwen2.5}, and reasoning variants OpenThinker-32b, and OpenThinker-7b \citep{OpenThoughts}; the reasoning variants are finetuned on math and coding task.
We prompt these models to interact with our environment, purely through natural language, without additional tools (see \autoref{fig:pseudo_code}, see \autoref{app:llm} for details).

\paragraph{Box's Apprentice.} The apprentice agent augments language models by enabling them to implement generative models of observed data. For model discovery, the agent writes a \texttt{pymc} program \citep{li2024automated} after 10 experiments, which is then fit and provided to the scientist explaining findings to the novice. For experimental design, the agent creates and uses these models to guide subsequent experiments.

% Box's Apprentice is an agent that augments the statistical modelling capabilities of the language model (\eg GPT-4o) by allowing it to write and fit a generative model to observed data and use them to make predictions \citep{li2024automated} (see \autoref{app:box} for implementation details). 
% This allows Box's Apprentice to build explicit representations of the underlying phenomena in each environment.
% In particular, for the model discovery experiments, we ask the LM agent to write a \texttt{pymc} program given the collected datapoints after 10 experiments. 
% We then fit this generative model and provide the implementation of the model and its inferred parameters to the scientist producing the explanation for the novice agent.
% For the experimental design experiments, we ask the LM agent to write a \texttt{pymc} program given the datapoints collected so far.
% The agent uses this model to guide its experiments.

\textbf{Experiment Setup.} For each environment, we run the agents for 5 independent trials. At each step, the agent chooses to perform an experiment, by specifying a design, and observes the outcome. After a fixed number of steps (0, 1, 3, 5, 7, 10), we evaluate the agent's performance using the metrics described earlier \autoref{sec:metrics}. The performance of models is averaged across 5 runs and over 10 evaluation points. 
We also explore a \textit{prior} vs \textit{no prior} condition  to investigate whether domain knowledge helps or hinders scientific discovery.
In the prior condition, we give the LM full context about the problem domain (\eg ``you are observing how participants balance delayed vs immediate rewards''), simulating scientists with background knowledge. 
In the no prior condition, we remove this context and describe the setting in a domain-agnostic way (\eg ``you receive a tuple of three values''), resembling reasoning from raw observations without preconceptions. This tests whether prior knowledge scaffolds discovery or creates biases that constrain exploration.

\subsection{Experimental Design Evaluation}

\begin{figure}
    \centering
    \includegraphics[width=\linewidth]{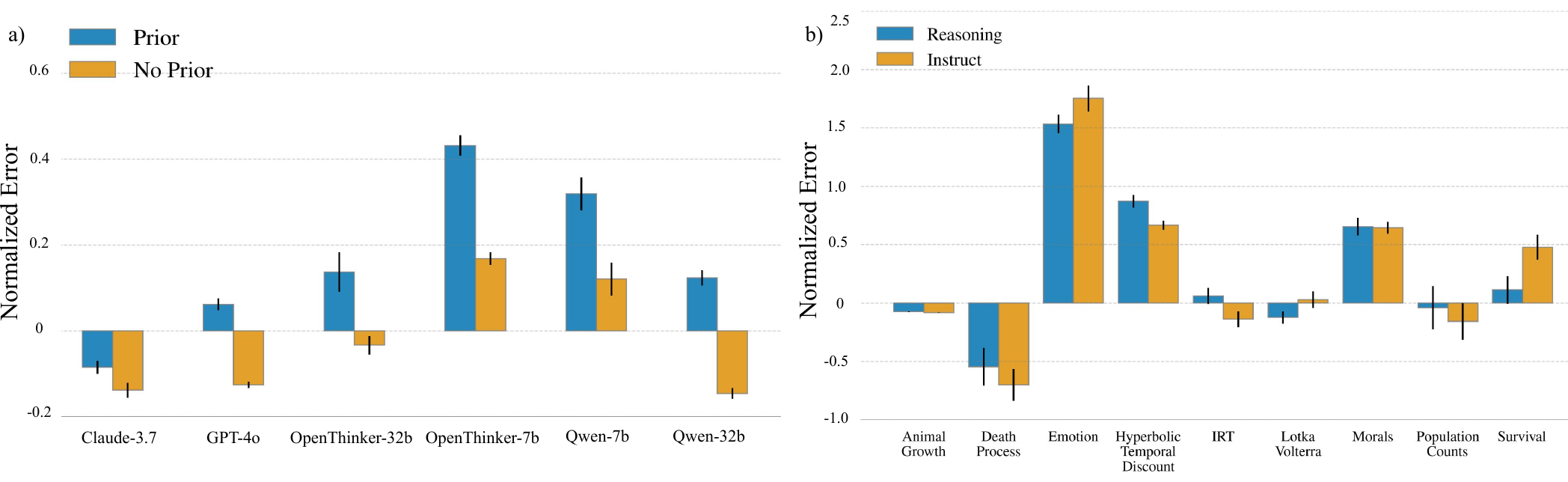}
    \caption{\textbf{Normalized Error Compared across Models.} (a) Comparison of the normalized errors for different LLMs with or without prior information included in the prompt. (b) Comparison of reasoning models (OpenThinker) and instruct models (Qwen) across environments. Error bars are the standard error across 5 runs.}
    \label{fig:oed-perf}
\end{figure}

\begin{figure}[t]
    \centering
\includegraphics[width=\textwidth]{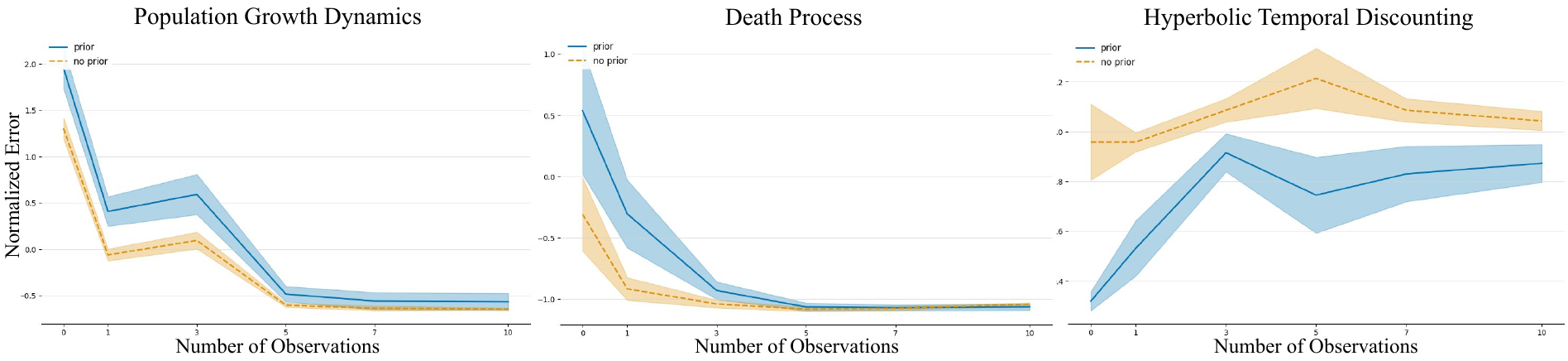}    \caption{\textbf{Normalized Errors Over Number of Observations.} Normalized errors for the LLM agent with \texttt{gpt-4o} with prior information (solid blue) and without prior information (dotted yellow) across three domains: Population Growth Dynamics (left), IRT (center) and Hyperbolic Discounting (right). Error bars are the standard error across 5 runs.}
    \label{fig:obs-oed}
    % \vspace{-5mm}
\end{figure}

\paragraph{Setup.} To evaluate the agents' performance, we first assess their ability to gather valuable information through their experiment selection and then measure how effectively they use this information to predict the environment. The Expected Information Regret (EI Regret) compares the Expected Information Gain (EIG) (\autoref{sec:eig}) of the agent's chosen experiments to the maximum EIG achievable from 100 random experiments. Lower EI Regret indicates more informative experiment selection. 
% To verify if agents can use observational data to make predictions, we measure their standardized prediction error about the environment before and after experimentation.

\paragraph{Prior information does not improve performance.} 
We find that models often perform better when given no prior information after 10 experiments (\autoref{fig:oed-perf}a). 
In some cases, this is because the LLM makes an overly strong assumption about the environment (\eg the signal decay is symmetric around the origin) and does not revise the assumption after more experiments; this is consistent with findings reported by \citet{li2024automated}.
In other cases, such as the hyperbolic discounting environment (\autoref{fig:obs-oed}, right), the model overfits to limited observations.

\paragraph{More experiments generally lead to better predictions.} 
We plot the learning trajectories for three environments in (\autoref{fig:obs-oed}). 
The agent's average prediction error decreases as it performs more experiments.
The Hyperbolic Temporal Discounting environments shows an unexpected trends where more experiments actually increases error. 
This may again be related to how prior knowledge interferes with effective learning from data.

\paragraph{Models Improve with Scale.} Larger models consistently outperform their smaller counterparts within the same model family. Both OpenThinker-32B and Qwen2.5-32B demonstrate significantly better performance than their respective 7B variants across environments (\autoref{fig:oed-perf}a), highlighting the benefits of scale for experimental design tasks.

\paragraph{Instruction-Tuned Models outperform Reasoning Models.} 
Surprisingly, the instruction-tuned Qwen2.5 models outperform the reasoning-focused OpenThinker models (\autoref{fig:oed-perf}b). This may be because OpenThinker models are finetuned to perform well on a relatively narrow set of verifiable problems in math and code, while instruction-tuned models retain broader capabilities that could be useful for experimental design.

\paragraph{Models performance varies substantially across environments.}
Models show varying performance across different environments (\autoref{fig:oed-perf}b). Performance is strongest on environments like population growth dynamics and death process, where the LM agent achieves negative standardized error, indicating that the LM successfully leveraged information gained through experimentation. However, in environments like hyperbolic discounting, performance is low even after experimentation, suggesting that some domains are inherently more challenging for current models.

\begin{figure}
    \centering
    \includegraphics[width=0.95\linewidth]{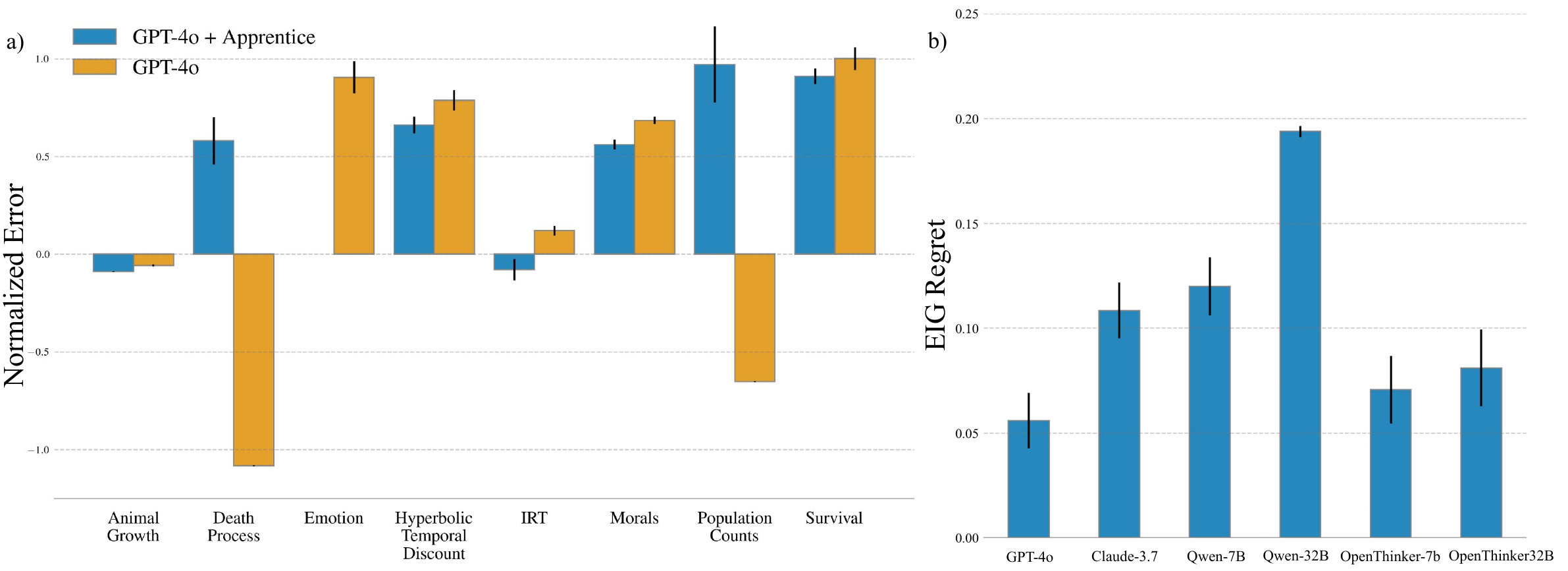}
    \caption{(a) Comparison of the Box's Apprentice with an LLM agent. (b) EIG Regret scores for six large language models, with lower values indicating better performance. }
    \label{fig:app-eig}
    \vspace{-3mm}
\end{figure}

\paragraph{EIG Regret reveals relationship between experimental design and prediction.}
Our EIG regret analysis (\autoref{fig:app-eig}b) provides insight into the relationship between two key components of scientific reasoning: designing informative experiments and making accurate predictions from collected data. GPT-4o achieves both the lowest EIG regret and strong predictive performance across several environments, suggesting these capabilities can be aligned. However, the varying performance of other models is informative --- for instance, Qwen-32B shows higher EIG regret despite good predictive performance in some domains, indicating that while these abilities may be related, excellence in prediction doesn't automatically translate to optimal experimental design.

\paragraph{LLMs cannot always optimally leverage statistical models.} While Box's Apprentice can propose and fit explicit statistical models to observed data, it does not consistently improve over the non-augmented LLM (GPT-4o) (\autoref{fig:app-eig}a)
From qualitative analysis of the models, we find that Box's Apprentice tends to favor overly simple functional forms due to limited data, such as using linear approximations for inherently nonlinear phenomena.

\subsection{Evaluating Model Discovery via Communication}
\begin{figure}
    \centering
    \includegraphics[width=\linewidth]{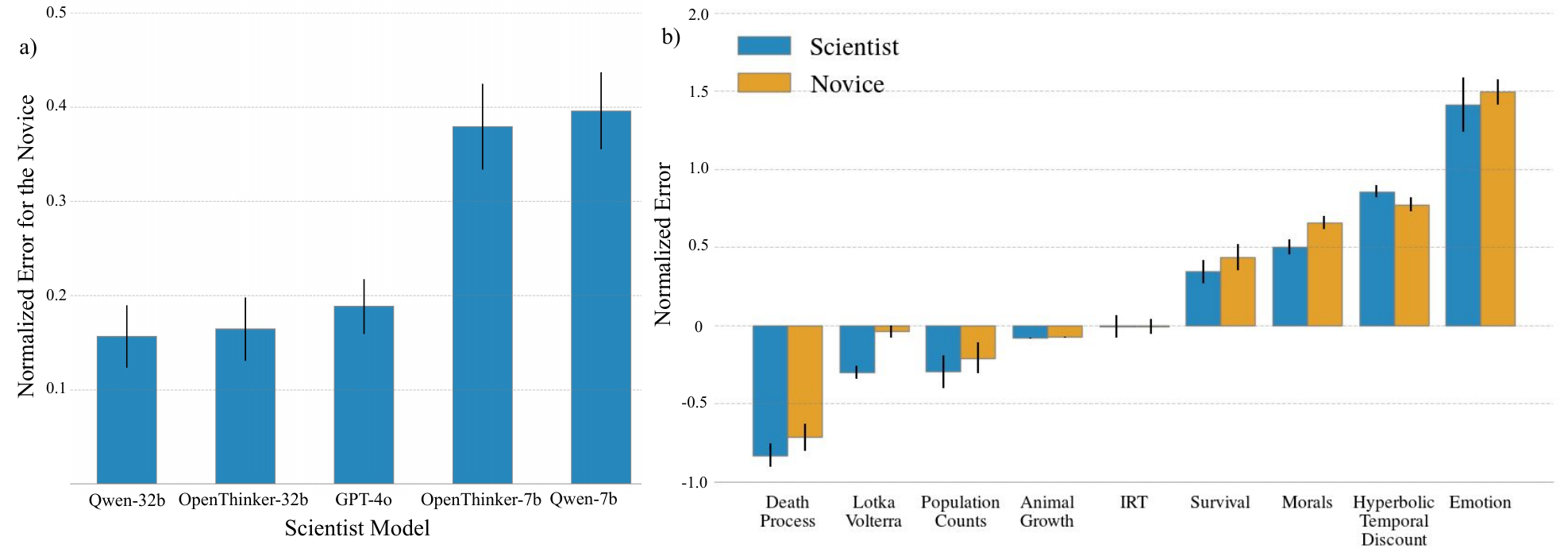}
    \vspace{-1mm}
    \caption{\textbf{Evaluation of Model Discovery via Communication.} (a) Comparison of the standardized error of the Novice (\texttt{gpt-4o}) with different Scientist models. (b) Comparison of errors made by the Novice and the Scientist (both models are \texttt{gpt-4o}). Error bars are standard error.
    }
    \label{fig:discovery}
    \vspace{-3mm}
\end{figure}

\paragraph{Setup.} Next, we evaluate the agents' ability to build and communicate models that capture the underlying phenomena in each environment. To test this, we have the agents interact with the environment for 10 steps (scientist phase) and then generate a natural language explanation of their findings. We then provide this explanation to a \textit{novice} agent, which must make predictions about the environment without any direct interaction (novice phase by using the explanation from the scientist;  \autoref{sec:comm}). 
The novice agent is always \texttt{gpt-4o}.
The scientist's prediction after 10 observations (Error After Experiments) acts as a weak positive control. Ideally, if the scientist's explanation is effective, the novice's error should approach the positive control.  

\vspace{-2mm}
\paragraph{Explanations improve with scale.} Larger models generally produce more effective explanations, as evidenced by better novice performance when using explanations from 32B variants compared to 7B models (\autoref{fig:discovery}a). This suggests that increased model scale improves not just experimentation but also the ability to distill and communicate findings.

\paragraph{Explanations are not as good as experiments.}
As expected, novice agents perform worse than scientists who directly interacted with the environment (\autoref{fig:discovery}b). 
The gap suggests that current explanation methods do not fully capture the knowledge gained through experimentation.

\paragraph{Explanations are more helpful for some environments.} However, the effectiveness of explanations varies substantially across domains (\autoref{fig:discovery}b). For instance, explanations are helpful for animal growth, but struggle with complex domains like moral judgments. This variation likely reflects the complexity of different domains and the current limitations of language models in capturing and communicating certain types of patterns.

\vspace{-3mm}
\section{Discussion}
\vspace{-3mm}
We introduced \bg, a benchmark measuring language-based agents' capabilities in experimental design and model discovery across 10 real-world-based environments. We evaluated experimental design using information gain metrics and developed a novel model discovery metric based on an agent's ability to explain its model to a novice agent.
Our evaluation across multiple model scales (7B-32B parameters) shows that while larger and closed-source models generally perform better, fundamental challenges persist. Neither domain-specific prior knowledge nor statistical modeling capabilities consistently improved performance. Some environments yielded strong results with larger models, while others remained challenging for all approaches.
\bg has limitations: it uses pre-defined experimental paradigms rather than requiring design from scratch \citep{dennis2020emergent}, ignores resource constraints, and covers limited scientific domains. Future work should address these limitations by incorporating experiment design from scratch, resource constraints, and more diverse fields \citep{huang2025automatedhypothesisvalidationagentic}. We could also expand the human behavior environments (Moral Machines, Emotions) with more sophisticated participant simulations \citep{ argyle2023out, aher2023using, shaikh2023rehearsal,park2022social, park2023generative}.
While our experiments demonstrated potential for interfaces that augment language models' scientific reasoning capabilities, future research should explore data visualization,  model validation \citep{li2024criticalcriticautomationlanguage}, and web-based research strategies to enhance experimental guidance and discovery.

\bibliography{main}
\bibliographystyle{plainnat}
\clearpage

\newpage
\section*{NeurIPS Paper Checklist}

\begin{enumerate}

\item {\bf Claims}
    \item[] Question: Do the main claims made in the abstract and introduction accurately reflect the paper's contributions and scope?
    \item[] Answer: \answerYes{} % Replace by \answerYes{}, \answerNo{}, or \answerNA{}.
    \item[] Justification: We describe the design of our benchmark accurately, summarize results with different models.
    \item[] Guidelines:
    \begin{itemize}
        \item The answer NA means that the abstract and introduction do not include the claims made in the paper.
        \item The abstract and/or introduction should clearly state the claims made, including the contributions made in the paper and important assumptions and limitations. A No or NA answer to this question will not be perceived well by the reviewers. 
        \item The claims made should match theoretical and experimental results, and reflect how much the results can be expected to generalize to other settings. 
        \item It is fine to include aspirational goals as motivation as long as it is clear that these goals are not attained by the paper. 
    \end{itemize}

\item {\bf Limitations}
    \item[] Question: Does the paper discuss the limitations of the work performed by the authors?
    \item[] Answer: \answerYes{}{} % Replace by \answerYes{}, \answerNo{}, or \answerNA{}.
    \item[] Justification: See discussion.
    \item[] Guidelines:
    \begin{itemize}
        \item The answer NA means that the paper has no limitation while the answer No means that the paper has limitations, but those are not discussed in the paper. 
        \item The authors are encouraged to create a separate "Limitations" section in their paper.
        \item The paper should point out any strong assumptions and how robust the results are to violations of these assumptions (e.g., independence assumptions, noiseless settings, model well-specification, asymptotic approximations only holding locally). The authors should reflect on how these assumptions might be violated in practice and what the implications would be.
        \item The authors should reflect on the scope of the claims made, e.g., if the approach was only tested on a few datasets or with a few runs. In general, empirical results often depend on implicit assumptions, which should be articulated.
        \item The authors should reflect on the factors that influence the performance of the approach. For example, a facial recognition algorithm may perform poorly when image resolution is low or images are taken in low lighting. Or a speech-to-text system might not be used reliably to provide closed captions for online lectures because it fails to handle technical jargon.
        \item The authors should discuss the computational efficiency of the proposed algorithms and how they scale with dataset size.
        \item If applicable, the authors should discuss possible limitations of their approach to address problems of privacy and fairness.
        \item While the authors might fear that complete honesty about limitations might be used by reviewers as grounds for rejection, a worse outcome might be that reviewers discover limitations that aren't acknowledged in the paper. The authors should use their best judgment and recognize that individual actions in favor of transparency play an important role in developing norms that preserve the integrity of the community. Reviewers will be specifically instructed to not penalize honesty concerning limitations.
    \end{itemize}

\item {\bf Theory assumptions and proofs}
    \item[] Question: For each theoretical result, does the paper provide the full set of assumptions and a complete (and correct) proof?
    \item[] Answer: \answerNA{} % Replace by \answerYes{}, \answerNo{}, or \answerNA{}.
    \item[] Justification: No proofs or new theoretical result.
    \item[] Guidelines:
    \begin{itemize}
        \item The answer NA means that the paper does not include theoretical results. 
        \item All the theorems, formulas, and proofs in the paper should be numbered and cross-referenced.
        \item All assumptions should be clearly stated or referenced in the statement of any theorems.
        \item The proofs can either appear in the main paper or the supplemental material, but if they appear in the supplemental material, the authors are encouraged to provide a short proof sketch to provide intuition. 
        \item Inversely, any informal proof provided in the core of the paper should be complemented by formal proofs provided in appendix or supplemental material.
        \item Theorems and Lemmas that the proof relies upon should be properly referenced. 
    \end{itemize}

    \item {\bf Experimental result reproducibility}
    \item[] Question: Does the paper fully disclose all the information needed to reproduce the main experimental results of the paper to the extent that it affects the main claims and/or conclusions of the paper (regardless of whether the code and data are provided or not)?
    \item[] Answer: \answerYes{} % Replace by \answerYes{}, \answerNo{}, or \answerNA{}.
    \item[] Justification: Yes, further, all our code, results and scripts are available on github.
    \item[] Guidelines:
    \begin{itemize}
        \item The answer NA means that the paper does not include experiments.
        \item If the paper includes experiments, a No answer to this question will not be perceived well by the reviewers: Making the paper reproducible is important, regardless of whether the code and data are provided or not.
        \item If the contribution is a dataset and/or model, the authors should describe the steps taken to make their results reproducible or verifiable. 
        \item Depending on the contribution, reproducibility can be accomplished in various ways. For example, if the contribution is a novel architecture, describing the architecture fully might suffice, or if the contribution is a specific model and empirical evaluation, it may be necessary to either make it possible for others to replicate the model with the same dataset, or provide access to the model. In general. releasing code and data is often one good way to accomplish this, but reproducibility can also be provided via detailed instructions for how to replicate the results, access to a hosted model (e.g., in the case of a large language model), releasing of a model checkpoint, or other means that are appropriate to the research performed.
        \item While NeurIPS does not require releasing code, the conference does require all submissions to provide some reasonable avenue for reproducibility, which may depend on the nature of the contribution. For example
        \begin{enumerate}
            \item If the contribution is primarily a new algorithm, the paper should make it clear how to reproduce that algorithm.
            \item If the contribution is primarily a new model architecture, the paper should describe the architecture clearly and fully.
            \item If the contribution is a new model (e.g., a large language model), then there should either be a way to access this model for reproducing the results or a way to reproduce the model (e.g., with an open-source dataset or instructions for how to construct the dataset).
            \item We recognize that reproducibility may be tricky in some cases, in which case authors are welcome to describe the particular way they provide for reproducibility. In the case of closed-source models, it may be that access to the model is limited in some way (e.g., to registered users), but it should be possible for other researchers to have some path to reproducing or verifying the results.
        \end{enumerate}
    \end{itemize}

\item {\bf Open access to data and code}
    \item[] Question: Does the paper provide open access to the data and code, with sufficient instructions to faithfully reproduce the main experimental results, as described in supplemental material?
    \item[] Answer: \answerYes{} % Replace by \answerYes{}, \answerNo{}, or \answerNA{}.
    \item[] Justification: All the code is accessible on the github.
    \item[] Guidelines:
    \begin{itemize}
        \item The answer NA means that paper does not include experiments requiring code.
        \item Please see the NeurIPS code and data submission guidelines (\url{https://nips.cc/public/guides/CodeSubmissionPolicy}) for more details.
        \item While we encourage the release of code and data, we understand that this might not be possible, so “No” is an acceptable answer. Papers cannot be rejected simply for not including code, unless this is central to the contribution (e.g., for a new open-source benchmark).
        \item The instructions should contain the exact command and environment needed to run to reproduce the results. See the NeurIPS code and data submission guidelines (\url{https://nips.cc/public/guides/CodeSubmissionPolicy}) for more details.
        \item The authors should provide instructions on data access and preparation, including how to access the raw data, preprocessed data, intermediate data, and generated data, etc.
        \item The authors should provide scripts to reproduce all experimental results for the new proposed method and baselines. If only a subset of experiments are reproducible, they should state which ones are omitted from the script and why.
        \item At submission time, to preserve anonymity, the authors should release anonymized versions (if applicable).
        \item Providing as much information as possible in supplemental material (appended to the paper) is recommended, but including URLs to data and code is permitted.
    \end{itemize}

\item {\bf Experimental setting/details}
    \item[] Question: Does the paper specify all the training and test details (e.g., data splits, hyperparameters, how they were chosen, type of optimizer, etc.) necessary to understand the results?
    \item[] Answer: \answerYes{} % Replace by \answerYes{}, \answerNo{}, or \answerNA{}.
    \item[] Justification: We describe this in detail in experimental setup and have the full specification in the appendix.
    \item[] Guidelines:
    \begin{itemize}
        \item The answer NA means that the paper does not include experiments.
        \item The experimental setting should be presented in the core of the paper to a level of detail that is necessary to appreciate the results and make sense of them.
        \item The full details can be provided either with the code, in appendix, or as supplemental material.
    \end{itemize}

\item {\bf Experiment statistical significance}
    \item[] Question: Does the paper report error bars suitably and correctly defined or other appropriate information about the statistical significance of the experiments?
    \item[] Answer: \answerYes{} % Replace by \answerYes{}, \answerNo{}, or \answerNA{}.
    \item[] Justification: We report statistical significance in all our results...
    \item[] Guidelines:
    \begin{itemize}
        \item The answer NA means that the paper does not include experiments.
        \item The authors should answer "Yes" if the results are accompanied by error bars, confidence intervals, or statistical significance tests, at least for the experiments that support the main claims of the paper.
        \item The factors of variability that the error bars are capturing should be clearly stated (for example, train/test split, initialization, random drawing of some parameter, or overall run with given experimental conditions).
        \item The method for calculating the error bars should be explained (closed form formula, call to a library function, bootstrap, etc.)
        \item The assumptions made should be given (e.g., Normally distributed errors).
        \item It should be clear whether the error bar is the standard deviation or the standard error of the mean.
        \item It is OK to report 1-sigma error bars, but one should state it. The authors should preferably report a 2-sigma error bar than state that they have a 96\% CI, if the hypothesis of Normality of errors is not verified.
        \item For asymmetric distributions, the authors should be careful not to show in tables or figures symmetric error bars that would yield results that are out of range (e.g. negative error rates).
        \item If error bars are reported in tables or plots, The authors should explain in the text how they were calculated and reference the corresponding figures or tables in the text.
    \end{itemize}

\item {\bf Experiments compute resources}
    \item[] Question: For each experiment, does the paper provide sufficient information on the computer resources (type of compute workers, memory, time of execution) needed to reproduce the experiments?
    \item[] Answer: \answerYes{} % Replace by \answerYes{}, \answerNo{}, or \answerNA{}.
    \item[] Justification: See appendix section B. 
    \item[] Guidelines:
    \begin{itemize}
        \item The answer NA means that the paper does not include experiments.
        \item The paper should indicate the type of compute workers CPU or GPU, internal cluster, or cloud provider, including relevant memory and storage.
        \item The paper should provide the amount of compute required for each of the individual experimental runs as well as estimate the total compute. 
        \item The paper should disclose whether the full research project required more compute than the experiments reported in the paper (e.g., preliminary or failed experiments that didn't make it into the paper). 
    \end{itemize}
    
\item {\bf Code of ethics}
    \item[] Question: Does the research conducted in the paper conform, in every respect, with the NeurIPS Code of Ethics \url{https://neurips.cc/public/EthicsGuidelines}?
    \item[] Answer: \answerYes{} % Replace by \answerYes{}, \answerNo{}, or \answerNA{}.
    \item[] Justification:  Single blind submission and we follow the code.
    \item[] Guidelines:
    \begin{itemize}
        \item The answer NA means that the authors have not reviewed the NeurIPS Code of Ethics.
        \item If the authors answer No, they should explain the special circumstances that require a deviation from the Code of Ethics.
        \item The authors should make sure to preserve anonymity (e.g., if there is a special consideration due to laws or regulations in their jurisdiction).
    \end{itemize}

\item {\bf Broader impacts}
    \item[] Question: Does the paper discuss both potential positive societal impacts and negative societal impacts of the work performed?
    \item[] Answer: \answerNo{} % Replace by \answerYes{}, \answerNo{}, or \answerNA{}.
    \item[] Justification: We don't discuss these as there are no direct negative societal impacts.
    \item[] Guidelines:
    \begin{itemize}
        \item The answer NA means that there is no societal impact of the work performed.
        \item If the authors answer NA or No, they should explain why their work has no societal impact or why the paper does not address societal impact.
        \item Examples of negative societal impacts include potential malicious or unintended uses (e.g., disinformation, generating fake profiles, surveillance), fairness considerations (e.g., deployment of technologies that could make decisions that unfairly impact specific groups), privacy considerations, and security considerations.
        \item The conference expects that many papers will be foundational research and not tied to particular applications, let alone deployments. However, if there is a direct path to any negative applications, the authors should point it out. For example, it is legitimate to point out that an improvement in the quality of generative models could be used to generate deepfakes for disinformation. On the other hand, it is not needed to point out that a generic algorithm for optimizing neural networks could enable people to train models that generate Deepfakes faster.
        \item The authors should consider possible harms that could arise when the technology is being used as intended and functioning correctly, harms that could arise when the technology is being used as intended but gives incorrect results, and harms following from (intentional or unintentional) misuse of the technology.
        \item If there are negative societal impacts, the authors could also discuss possible mitigation strategies (e.g., gated release of models, providing defenses in addition to attacks, mechanisms for monitoring misuse, mechanisms to monitor how a system learns from feedback over time, improving the efficiency and accessibility of ML).
    \end{itemize}
    
\item {\bf Safeguards}
    \item[] Question: Does the paper describe safeguards that have been put in place for responsible release of data or models that have a high risk for misuse (e.g., pretrained language models, image generators, or scraped datasets)?
    \item[] Answer: \answerNA{} % Replace by \answerYes{}, \answerNo{}, or \answerNA{}.
    \item[] Justification: Not relevant for the paper.
    \item[] Guidelines:
    \begin{itemize}
        \item The answer NA means that the paper poses no such risks.
        \item Released models that have a high risk for misuse or dual-use should be released with necessary safeguards to allow for controlled use of the model, for example by requiring that users adhere to usage guidelines or restrictions to access the model or implementing safety filters. 
        \item Datasets that have been scraped from the Internet could pose safety risks. The authors should describe how they avoided releasing unsafe images.
        \item We recognize that providing effective safeguards is challenging, and many papers do not require this, but we encourage authors to take this into account and make a best faith effort.
    \end{itemize}

\item {\bf Licenses for existing assets}
    \item[] Question: Are the creators or original owners of assets (e.g., code, data, models), used in the paper, properly credited and are the license and terms of use explicitly mentioned and properly respected?
    \item[] Answer: \answerYes{} % Replace by \answerYes{}, \answerNo{}, or \answerNA{}.
    \item[] Justification: All models have been cited appropriately. The papers that inspired the environments have been credited too.
    \item[] Guidelines:
    \begin{itemize}
        \item The answer NA means that the paper does not use existing assets.
        \item The authors should cite the original paper that produced the code package or dataset.
        \item The authors should state which version of the asset is used and, if possible, include a URL.
        \item The name of the license (e.g., CC-BY 4.0) should be included for each asset.
        \item For scraped data from a particular source (e.g., website), the copyright and terms of service of that source should be provided.
        \item If assets are released, the license, copyright information, and terms of use in the package should be provided. For popular datasets, \url{paperswithcode.com/datasets} has curated licenses for some datasets. Their licensing guide can help determine the license of a dataset.
        \item For existing datasets that are re-packaged, both the original license and the license of the derived asset (if it has changed) should be provided.
        \item If this information is not available online, the authors are encouraged to reach out to the asset's creators.
    \end{itemize}

\item {\bf New assets}
    \item[] Question: Are new assets introduced in the paper well documented and is the documentation provided alongside the assets?
    \item[] Answer: \answerYes{} % Replace by \answerYes{}, \answerNo{}, or \answerNA{}.
    \item[] Justification: We add documentation to the \bg code.
    \item[] Guidelines:
    \begin{itemize}
        \item The answer NA means that the paper does not release new assets.
        \item Researchers should communicate the details of the dataset/code/model as part of their submissions via structured templates. This includes details about training, license, limitations, etc. 
        \item The paper should discuss whether and how consent was obtained from people whose asset is used.
        \item At submission time, remember to anonymize your assets (if applicable). You can either create an anonymized URL or include an anonymized zip file.
    \end{itemize}

\item {\bf Crowdsourcing and research with human subjects}
    \item[] Question: For crowdsourcing experiments and research with human subjects, does the paper include the full text of instructions given to participants and screenshots, if applicable, as well as details about compensation (if any)? 
    \item[] Answer: \answerNA{} % Replace by \answerYes{}, \answerNo{}, or \answerNA{}.
    \item[] Justification: No human participants were recruited.
    \item[] Guidelines:
    \begin{itemize}
        \item The answer NA means that the paper does not involve crowdsourcing nor research with human subjects.
        \item Including this information in the supplemental material is fine, but if the main contribution of the paper involves human subjects, then as much detail as possible should be included in the main paper. 
        \item According to the NeurIPS Code of Ethics, workers involved in data collection, curation, or other labor should be paid at least the minimum wage in the country of the data collector. 
    \end{itemize}

\item {\bf Institutional review board (IRB) approvals or equivalent for research with human subjects}
    \item[] Question: Does the paper describe potential risks incurred by study participants, whether such risks were disclosed to the subjects, and whether Institutional Review Board (IRB) approvals (or an equivalent approval/review based on the requirements of your country or institution) were obtained?
    \item[] Answer: \answerNA{} % Replace by \answerYes{}, \answerNo{}, or \answerNA{}.
    \item[] Justification: Paper does not use human participants.
    \item[] Guidelines:
    \begin{itemize}
        \item The answer NA means that the paper does not involve crowdsourcing nor research with human subjects.
        \item Depending on the country in which research is conducted, IRB approval (or equivalent) may be required for any human subjects research. If you obtained IRB approval, you should clearly state this in the paper. 
        \item We recognize that the procedures for this may vary significantly between institutions and locations, and we expect authors to adhere to the NeurIPS Code of Ethics and the guidelines for their institution. 
        \item For initial submissions, do not include any information that would break anonymity (if applicable), such as the institution conducting the review.
    \end{itemize}

\item {\bf Declaration of LLM usage}
    \item[] Question: Does the paper describe the usage of LLMs if it is an important, original, or non-standard component of the core methods in this research? Note that if the LLM is used only for writing, editing, or formatting purposes and does not impact the core methodology, scientific rigorousness, or originality of the research, declaration is not required.
    %this research? 
    \item[] Answer: \answerNA{} % Replace by \answerYes{}, \answerNo{}, or \answerNA{}.
    \item[] Justification: None of the core methods used LLMs.
    \item[] Guidelines:
    \begin{itemize}
        \item The answer NA means that the core method development in this research does not involve LLMs as any important, original, or non-standard components.
        \item Please refer to our LLM policy (\url{https://neurips.cc/Conferences/2025/LLM}) for what should or should not be described.
    \end{itemize}

\end{enumerate}

\clearpage
\appendix

\newpage
\section{Full Results}
See \autoref{apptab:res1}, and \autoref{apptab:res2} for \footnote{We omit the predatory-prey and Emotions domains for Box's Apprentice, since GPT-4o could not reliably produce \texttt{pymc} programs} prediction errors across all environments for GPT-4o and the Box's apprentice with GPT-4o. Full results are available in the Github Repository.

\begin{table}[tbp]
% \centering
\begin{tabularx}{\textwidth}{ccccc}
\toprule
\textbf{Env} & \textbf{Goal} & \textbf{Error@0} & \textbf{Error@10} & \textbf{Discovery@10} \\
\midrule
Hyperbolic Discounting & Choice & \makecell{0.32$\pm$0.04 \\ 0.96$\pm$0.15} & \makecell{0.87$\pm$0.08 \\ 1.04$\pm$0.04} & \makecell{0.79$\pm$0.37 \\ 0.96$\pm$0.07} \\
\midrule
Hyperbolic Discounting  & Discount & \makecell{-0.06$\pm$0.00 \\ -} & \makecell{-0.06$\pm$0.00 \\ -} & \makecell{-\\ -} \\
\midrule
Location Finding & Signal & \makecell{0.30$\pm$0.25 \\ 0.63$\pm$0.39} & \makecell{0.59$\pm$0.55 \\ 0.86$\pm$0.47}  & \makecell{4.75$\pm$4.51 \\ 1.52$\pm$1.28} \\
\midrule
Location Finding & Source Location & \makecell{1.29$\pm$1.3 \\ -} & \makecell{-0.15 $\pm$0.4 \\ -}  & \makecell{- \\ -} \\
\midrule
Death Process & Num Infected & \makecell{0.54$\pm$0.52 \\ -0.31$\pm$0.30} & \makecell{-1.06$\pm$0.03 \\ -1.04$\pm$0.01} & \makecell{-1.08$\pm$0.01 \\ -1.00$\pm$0.11} \\
\midrule
Death Process  & Infection Rate & \makecell{0.13$\pm$0.37 \\ -} & \makecell{1.64$\pm$1.12 \\ -} &  \makecell{- \\ -} \\
\midrule
IRT & Correctness & \makecell{0.12$\pm$0.07 \\ 0.08$\pm$0.15} & \makecell{-0.24$\pm$0.10 \\ 0.00$\pm$0.13} &  \makecell{0.12$\pm$0.18 \\ 0.12$\pm$0.14} \\
\midrule
Dugongs & Length & \makecell{-0.04$\pm$0.02 \\ -0.04$\pm$0.02} & \makecell{-0.08$\pm$0.00 \\ -0.08$\pm$0.00} &  \makecell{-0.06$\pm$0.04 \\ -0.07$\pm$0.02} \\
\midrule
Peregrines & Population & \makecell{1.95$\pm$0.22 \\ 1.30$\pm$0.11} & \makecell{-0.57$\pm$0.09 \\ -0.65$\pm$0.01} &  \makecell{-0.65$\pm$0.02 \\ -0.66$\pm$0.03} \\
\midrule
Mastectomy & Survival & \makecell{0.04$\pm$0.14 \\ 0.32$\pm$0.08} & \makecell{0.36$\pm$0.10 \\ 0.27$\pm$0.12} &  \makecell{1.00$\pm$0.41 \\ 0.45$\pm$0.18} \\
\midrule
Predator-Prey & Population & \makecell{0.38$\pm$0.04 \\ 0.75$\pm$0.02} & \makecell{-0.31$\pm$0.05 \\ -0.42$\pm$0.01} & \makecell{-0.01$\pm$0.12 \\ -0.07$\pm$0.40} \\
\midrule
Emotions & Prediction & \makecell{1.04$\pm$0.21 \\ N/A} & \makecell{1.22$\pm$0.29 \\ N/A} & \makecell{0.90$\pm$0.58 \\ N/A} \\
\midrule
Moral Machines & Judgement & \makecell{0.40$\pm$0.07 \\ N/A} & \makecell{0.36$\pm$0.04 \\ N/A}  & \makecell{0.68$\pm$0.13 \\ N/A} \\
\bottomrule
\end{tabularx}
\vspace{2mm}
\caption{\textbf{Performance of GPT-4o Across Different Tasks.} Numbers shown are normalized-0 errors. Errors with prior (top line) and without prior (bottom line) appear on different lines. Errors are averaged across 5 runs.}
\label{apptab:res1}
\vspace{-3mm}
\end{table}

\begin{table}[tbp]
\centering
\begin{tabularx}{\textwidth}{ccccc}
\toprule
\textbf{Env} & \textbf{Goal} & \textbf{Error@0} & \textbf{Error@10} & \textbf{Discovery@10} \\
\midrule
Hyperbolic Discounting & Choice & \makecell{0.66 $\pm$ 0.25\\0.66 $\pm$ 0.25} & \makecell{1.17 $\pm$ 0.14 \\ 0.91 $\pm$ 0.09} & \makecell{0.66 $\pm$ 0.30 \\ 0.74 $\pm$ 0.42} \\
\midrule
Location Finding & Signal & \makecell{0.99 $\pm$ 0.58  \\ 1.18 $\pm$ 0.64} & \makecell{1.45 $\pm$ 1.60 \\ 0.83 $\pm$ 0.60} &  \makecell{1.18 $\pm$ 1.12 \\ -0.01 $\pm$ 0.30} \\
 \midrule
Death Process & Num Infected &
\makecell{3.79 $\pm$ 1.68 \\ -0.90 $\pm$ 0.05} &  \makecell{-1.02 $\pm$ 0.05 \\ -0.61 $\pm$ 0.30} &  \makecell{0.58 $\pm$ 0.85 \\ 0.50 $\pm$ 1.26}  \\
 \midrule
% IRT & Correctness & \makecell{0.44 $\pm$ 0.36 \\ 0.12} & \makecell{-0.12 $\pm$ 0.14\\ 0.12 $\pm$ 0.14}  & \makecell{-0.08 $\pm$ 0.39 \\ 0.2 $\pm$ 0.40} \\
IRT & Correctness & \makecell{0.44 $\pm$ 0.36 \\ $0.12 \pm 0.24$ } & \makecell{$-0.12 \pm 0.14$ \\ $0.12 \pm 0.14$} & \makecell{-0.08 $\pm$ 0.39 \\ 0.2 $\pm$ 0.40} \\
\midrule
Dugongs & Length & \makecell{0.26 $\pm$ 0.12 \\ 0.05 $\pm$ 0.10} & \makecell{-0.08 $\pm$ 0.02 \\ -0.09 $\pm$ 0.004} & \makecell{$-0.09 \pm 0.005$ \\ $-0.08 \pm 0.004$} \\
\midrule
Peregrines & Population & \makecell{2.71 $\pm$ 0.60 \\ 1.62 $\pm$ 0.47} & \makecell{0.04 $\pm$ 0.21 \\  0.95 $\pm$ 0.86} & \makecell{0.97 $\pm$ 1.38 \\  -0.19 $\pm$ 0.79} \\
\midrule
Mastectomy & Survival & \makecell{0.14 $\pm$ 0.41 \\ 0.73 $\pm$ 0.15} & \makecell{0.55 $\pm$ 0.24 \\ 0.64 $\pm$ 0.15} & \makecell{0.91 $\pm$ 0.28 \\ 0.27 $\pm$ 0.23} \\
\midrule
Moral Machines & Judgement & \makecell{0.97 $\pm$ 0.33}  & \makecell{0.89 $\pm$ 0.21} & \makecell{0.56 $\pm$ 0.18}  \\
\bottomrule
\end{tabularx}
\vspace{1mm}
\caption{\textbf{Performance of Box's Apprentice Across Different Tasks.} Standardized errors shown here. Errors with prior (top line) and without prior (bottom line) appear on different lines. Errors are averaged across 5 runs.}
\label{apptab:res2}
\figurespace{}
\end{table}

\section{LLM Agent} \label{app:llm}
The LLM agent provides an easy way for a large language model (LLM) to interact with \bg.
By tailoring the system message to the specific environment, we can clearly define goals for the LLM, elicit experimental designs from it, make accurate predictions for queries, and generate explanations for a novice.
This agent class also incorporates a simple retry mechanism that allows the LLM to correct its designs if they are initially invalid.

Models were configured with a temperature parameter of 0.0 to ensure deterministic outputs. Maximum token limits were set to 512 tokens for instruct models and 1024 tokens for thinking variants, providing sufficient thinking tokens for generating an answer without multiple retries.

GPT-4o and Claude-3.7-Sonnet were accessed via their APIs, while all other models were deployed using vLLM. For the vLLM-served models, we utilized a dual A40 GPU configuration: one GPU dedicated to model serving and the other for inference execution through the vLLM endpoint. This architecture ensured optimal resource allocation and performance stability throughout the experimental process.

Each OED experimental run consisted of 10 predictions conducted after 0, 1, 3, 5, 7, and 10 observations, respectively. Comprehensive log files were generated for each set of predictions to facilitate subsequent analysis. Execution time varied across model architectures, with most configurations requiring approximately 2-3 minutes per run (defined as a single seed, configuration, and environment combination). Models accessed through external APIs typically required longer execution times due to network latency and rate limiting considerations. Discovery experiments reduced execution times compared to OED experiments due to the decreased number of required API calls.

\section{Box's Apprentice} \label{app:box}
We closely follow \citet{li2024automated}.
In particular, to generate a candidate, we sample a single probabilistic program $z$ from the proposal LM, $q_{\text{LM}}(\cdot)$.
For the model discovery experiments, we perform this once after 10 experiments. 
For the OED experiments, we perform this three times over the course of 10 experiments. 
In all experiments, we use \texttt{GPT-4o} (\texttt{gpt-4o-2024-05-13}). 
The proposal LM $q_{\text{LM}}$ 
``conditions'' on  $h^{t} \in \Sigma^{*}$, a natural language instruction synthesizing previous modeling approaches and suggesting new approaches, the previous program $z^{t-1}$, and a textual representation of the dataset $\mathcal{D}$. 
\begin{align*}    
    z^t \sim q_{\text{LM}}(\cdot | z^{t-1}, h^{t-1}, \mathcal{D}).
\end{align*}
We run this at a temperature of 0.0.
Chain-of-thought reasoning, or generating intermediate reasoning steps, improves the performance of LMs \citep{wei2022chain}.
Motivated by this, we instruct $q_{\text{LM}}$
to reflect on the properties of the dataset, sketch a high-level modeling approach, state the hypotheses that it will address before writing a program, and add comments to code.
See the system prompt in Figure~\ref{code:box_system_prompt}.

\newpage
\begin{figure*}[htpb]
\centering
\begin{tcolorbox}[
width=1\textwidth,
title={Box's Apprentice system prompt}]
\fontsize{5pt}{5pt}\selectfont
\ttfamily
\begin{lstlisting}[language={}]
You are a brilliant statistician modeling a dataset.
Your job is to come up with a generative model that explains the true data by writing a pymc probabilistic program. 
Here is a description of the dataset {dataset_description}
{dataset_text_representation}
Here is a description of the columns {column_description}
If you are in the first round, you will not receive any additional information.
However, for the second round and beyond, I will give you the model you proposed previously.
Please import pymc NOT pymc3!
Note that there are differences in the arguments pymc expects. 
IMPORTANT: do not use sd as an argument use sigma instead!
It is crucial that you pass the idata_kwargs argument to pm.sample!!
IMPORTANT: Use the variable name "y_obs" for the observations when you define it!
IMPORTANT: Use the variable name "y_obs" for the observations when you define it!
IMPORTANT: Index the appropriate column names when grabbing data from observed_data. These column names are indicated in the column description.

Your answer should follow the template in the following order.
1. First, sketch a high-level probabilistic program for the data.
  You will go through multiple rounds of revision. 
  If there's a previous program in your context window and a list of hypotheses, revise based on this information!
  Explicitly cite the hypotheses (if there are any) that you address in your sketch.
2. After coming up with a plan, write your program and add comments to lines of code that address certain hypotheses.
```python
  import pymc as pm
  import numpy as np
  def gen_model(observed_data):
      # convert observed_data columns to numpy arrays
      # index the appropriate column names
      
      ....
      rng1 = np.random.default_rng(42)
      rng2 = np.random.default_rng(314)
      with pm.Model as model():
          # create a pm.MutableData object for each non-observation column 
          ...Your code here...
          # Copy the rest of this code verbatim but remember to have this indented in scope of model()!
          trace = pm.sample(1000, tune=500, target_accept=0.90, chains=3, cores=1, random_seed=rng1)
          posterior_predictive = pm.sample_posterior_predictive(trace, random_seed=rng2, return_inferencedata=False)
          return model, posterior_predictive, trace
```
\end{lstlisting}
\end{tcolorbox}
\caption{\textbf{\texttt{BoxLM} system prompt}
The system prompt for the proposal $p_{LM}$. We also include some additional instructions on \texttt{pymc} syntax such as wrapping features in a MutableData container. 
}
\label{code:box_system_prompt}
\end{figure*}

\section{Domains} \label{app:domains}

\subsection{Location Finding} 

The location finding environment has hidden signal sources that emit a signal. The scientist can makeg measurements of the superimposed signal at various points. The experiment is directly taken from \citet{foster19}. In table \ref{table:location-finding-experiment}, we describe the inputs and outputs of the experiment.

\begin{table}[ht]
\centering
\begin{tabular}{ll}
\toprule
\textbf{Parameter} & \textbf{Description} \\
\midrule
Model & Superposition of $K$ signal sources in $d$-dim space \\
Setup Parameters & Num signal sources $K$, dim of space $d$, base signal $b$, max signal $m$, noise $\sigma$\\
Observations & Total noisy signal at point of measurement \\
Goals & Predicting signal intensity at new points and source locations\\
\bottomrule
\end{tabular}
\vspace{2mm}
\caption{Location Finding}
\label{table:location-finding-experiment}
\end{table}

We define $k=3$ signal sources in $\mathbb{R}^d=\mathbb{R}^2$ space with locations at $\theta_k$. The number of sources is predefined and is known to the agent. Each source emits a signal strength $\alpha_k$. In our implementation, we choose $\alpha_k$ to be fixed for all sources. The signal strength decays according to the inverse square law--if an agent measures at point $\xi$, then the noisy superimposed signal observed will be distributed according to $\mathcal{N}(\mu(\theta,\xi),\sigma)$ where $\sigma$ is the signal noise, $\mu(\theta,\xi)$ is the total intensity at point $\xi$,

\begin{equation}
\mu(\theta,\xi)=b+\sum_{k=1}^K\frac{\alpha_k}{m+\mid\mid \theta_k-\xi \mid\mid^2}
\end{equation}

and $b,m>0$ are constants governing background and maximum signal. Note that unlike \citet{foster21a}, we observe the total intensity, not the log total intensity.

\subsection{Hyperbolic Discounting}
\label{htd}
The hyperbolic discounting domain has two hidden variables $(k,\alpha)$ to describe a participant's behavior, where each participant is asked to choose between an immediate reward $\$iR$ or a delayed reward $\$dR$ in $D$ days. The experiment is outlined in table \ref{table:hyperbolic-temporal-discount-experiment} below. 

\begin{table}[ht]
\centering
\begin{tabular}{ll}
\toprule
\textbf{Parameter} & \textbf{Description} \\
\midrule
Model & Human decision-making in temporal discounting of rewards \\
Setup Parameters & Params of the discount function ($\epsilon$, mean and std for $\log k$, scale for $\alpha$) \\
Observations & Choice between immediate $iR$ and delayed reward $dR$ at delay $D$ \\
Goals & Predicting choices and the value of the discount factor\\
\bottomrule
\end{tabular}
\vspace{2mm}
\caption{Hyperbolic Discounting}
\label{table:hyperbolic-temporal-discount-experiment}
\end{table}

In each measurement, we require $iR$ is strictly smaller than $dR$ and all three values have to be positive, because we assume a rational participant would always choose a higher immediate reward over a lower delayed reward. We follow the prior distribution of the latent variables given by  \citet{foster19}: 

\begin{equation}
\log k \sim N(-4.25, 1.5), \alpha \sim HalfNormal(0,2)
\end{equation}

where the HalfNormal distribution is a normal distribution truncated at 0. For each test, there are three variables in design: $iR$, $dR$, and $D$. We give values to each choice: receiving the immediate reward $\$iR$ has value $V_i = iR$, while receiving the delayed reward $\$dR$ in $D$ days has value $V_d = \frac{dR}{1+kD}$. Then, whether each participant's chooses the delayed reward in each scenario is characterized as a Bernoulli random variable $X\sim Bernoulli(p)$ where the probability of choosing the delayed reward is given by

\begin{equation}
p(X=1 | k,\alpha, iR, dR, D) = \epsilon + (1-2\epsilon) \Phi (\frac{V_d-V_i}{\alpha})
\end{equation}

where $\Phi$ is the cumulative distribution function of the standard normal distribution. In our implementation, we set $\epsilon = 0.01$ for all scenarios.

\subsection{Death Process}
\label{death process}
The death process environment models an infection spreading among a healthy population of $N$ individuals. The infection rate $\theta$ determines how the probability of infection increases over time. The environment is outlined in table \ref{table:death-process-experiment} below. 

\begin{table}[ht]
\centering
\begin{tabular}{ll}
\toprule
\textbf{Parameter} & \textbf{Description} \\
\midrule
Model & The spread of an infection over time \\
Setup Parameters & Pop size $N$, params of the infetion rate ($\mu$, $\sigma$, upper and lower bounds) \\
Observations & Number of infected individuals at observation time \\
Goals & Predicting the number of infected individuals at a time and the infection rate\\
\bottomrule
\end{tabular}
\vspace{2mm}
\caption{Death Process}
\label{table:death-process-experiment}
\end{table}

In our model, $\theta$ is given by the prior distribution outlined in \citet{foster21a}.

\begin{equation}
\theta \sim \text{TruncatedNormal}(\mu = 1, \sigma = 1, min = 0, max = \infty)
\end{equation}

The number of infected individuals $Y$ at time $t$ is distributed as a binomial random variable:

\begin{equation}
    Y|\theta, t\sim \text{Binomial}(N, \eta)
\end{equation}

where $\eta = 1-e^{-\theta t}$, and $N$ is the population size. We ask the agent to make observations sequentially by giving a time $t>0$ at each step.

\subsection{IRT}
\paragraph{1PL IRT Model}
The one parameter IRT (or Rasch) domain models the performance of multiple students on multi-question exams. The binary outcome (whether the student is correct) of a student-question pair is determined by latent variables governing the student's proficiency and the question's difficulty (Figure 2). The agent's goal is to predict the outcome of a particular student-question pair. The agent may observe other student-question pairs to view their outcome. Table \ref{table:1pl-irt-model-experiment} below details the inputs, outputs, and target for every variation of the IRT model.

\begin{table}[h]
\centering
\begin{tabular}{ll}
\toprule
\textbf{Param} & \textbf{Description} \\
\midrule
Model & Student performance on multi-question exams \\
Setup Parameters & Number of students $N$, number of questions $Q$, student-question pair to predict \\
Observations & Outcomes of various student-question pairs \\
Goals & Predicting the correctness of student responses to questions \\
\bottomrule
\end{tabular}
\vspace{2mm}
\caption{IRT Model}
\label{table:1pl-irt-model-experiment}
\end{table}

We define the ability $\alpha_j$ of student $j$ and the difficulty $\beta_k$ of question $k$. In our implementation, $\alpha$ and $\beta$ are standard normals. The outcome $O_{jk}$ of a student $j$ on question $k$ is determined by a Bernoulli trial where the probability of success $p_{jk}$ is determined by the logit function of $z_{jk}=\alpha_j-\beta_k$.

\begin{equation}
p_{jk}=\frac{1}{1+e^{-z_{jk}}}
\end{equation}

In summary, for a given student-question pair, we compute the probability of the student getting the question correct and return the result of the corresponding Bernoulli trial.

\paragraph{2PL IRT Model}
The two parameter IRT model is identical to the 1PL variant with an additional variable governing the discriminability $\gamma_k$ of question $k$. The discriminability models how sensitive the question is to incorrect answers. For higher values of $\gamma$, the probability of a student's answer being correct is higher. Thus the outcome $O_{jk}$ of a student $j$ on question $k$ is determined by a Bernoulli trial where the probability of success $p_{jk}$ is determined by the logit function of $z_{jk}=\gamma_k(\alpha_j-\beta_k)$.
\paragraph{3PL IRT Model}
The three parameter IRT model is identical to the 2PL variant with an additional variable modeling how susceptible a question is to guessing. For question $k$, $c_k$ determines the probability that a student gets the question right by guessing. Thus the outcome $O_{jk}$ of a student $j$ on question $k$ is determined by a Bernoulli trial where the probability of success $p_{jk}$ is determined by 

\begin{equation}
    p_{jk}=c_k+(1-c_k)\frac{1}{1+e^{-z_{jk}}}
\end{equation}

where $z_{jk}=\gamma_k(\alpha_j-\beta_k)$ as in 2PL.

We use the 2PL model in \bg.

\subsection{Dugongs}
The dugongs environment has the ages and lengths of dugongs (sea cows)\citep{Magnusson_posteriordb_a_set_2023}. The goal is to model the length of a dugong based on its age. The following table describes the inputs and outputs of the experiment:

\begin{table}[h]
\centering
\begin{tabular}{ll}
\toprule
\textbf{Parameter} & \textbf{Description} \\
\midrule
Model & Bayesian hierarchical model \\
Setup Parameters & alpha, beta, lambda, lower limit, upper limit \\
Observations & Length of dugong at a given age \\
Goals & Predicting the length of dugongs at different ages \\
\bottomrule
\end{tabular}
\vspace{2mm}
\caption{Dugongs Environment}
\label{table:dugong-experiment}
\end{table}

In this environment, the length of a dugong at age $x$ is modeled using a hierarchical Bayesian model with parameters $\alpha$, $\beta$, and $\lambda$. The age values range between 0 and 5. The observed length $Y$ at a given age $x$ is generated from a normal distribution with a mean that is a function of $x$ and the parameters $\alpha$, $\beta$, and $\lambda$, and a fixed standard deviation. The function representing the mean length $m$ is defined as:

\begin{equation}
m = \alpha - \beta \cdot |\lambda|^x
\end{equation}

The observed lengths are then drawn from a normal distribution:

\begin{equation}
Y \sim \mathcal{N}(m, \sigma)
\end{equation}

where $\sigma$ is the noise in the observed lengths, set to a fixed value (e.g., 0.25).

\subsection{Peregrines}
The peregrine environment models the population count of peregrine falcons at different times \citep{Magnusson_posteriordb_a_set_2023}. The goal is to understand how the population changes over time. The following table describes the inputs and outputs of the experiment:

\begin{table}[h]
\centering
\begin{tabular}{ll}
\toprule
\textbf{Parameter} & \textbf{Description} \\
\midrule
Model & Poisson regression model \\
Setup Parameters & Regression params: $\alpha$, $\beta_1$, $\beta_2$, and $\beta_3$ \\
Observations & Population count of peregrine falcons at a given time \\
Goals & Predicting the population of peregrines at different times \\
\bottomrule
\end{tabular}
\vspace{2mm}
\caption{Peregrine Environment}
\label{table:peregrine-experiment}
\end{table}

In this environment, the population count of peregrine falcons at time $t$ is modeled using a Poisson regression model with parameters $\alpha$, $\beta_1$, $\beta_2$, and $\beta_3$ . The time values range between 0 and 5. The population count $C$ at a given time $t$ is generated from a Poisson distribution with a mean that is a function of $t$ and the parameters $\alpha$, $\beta_1$, $\beta_2$, and $\beta_3$. The function representing the log of the mean population count $\lambda$ is defined as:

\begin{equation}
\log \lambda = \alpha + \beta_1 t + \beta_2 t^2 + \beta_3 t^3
\end{equation}

The observed population counts are then drawn from a Poisson distribution:

\begin{equation}
C \sim \text{Poisson}(\exp(\log \lambda))
\end{equation}

This model allows for capturing the non-linear trends in the population data over time.

\subsection{Survival Analysis: Mastectomy}

The survival analysis environment models the outcomes of breast cancer patients based on the time since surgery and the metastasized status. The following table describes the inputs and outputs of the experiment:

\begin{table}[ht]
\centering
\begin{tabular}{ll}
\toprule
\textbf{Parameter} & \textbf{Description} \\
\midrule
Model & Survival analysis using a Bayesian approach \\
Setup Parameters & num\_patients, time\_upper\_bound, lambda, beta \\
Observations & Whether a selected patient is alive or dead \\
Goals & Predict survival based on time since surgery and if the cancer had metastasized\\
\bottomrule
\end{tabular}
\vspace{2mm}
\caption{Survival Analysis Environment}
\label{table:survival-analysis-experiment}
\end{table}

In this environment, the outcome (alive or dead) of a patient is modeled based on the time since surgery and whether the cancer metastasized \citep{cox2018analysis}. The outcomes are generated using a Bayesian model with parameters $\lambda_0$ and $\beta$. The number of patients and the upper bound of the time since surgery are configurable. At the start of an episode, we sample a set of patients that have undergone mastectomy, with varying times since they had surgery and if their cancer had metastasized or not. The experimenter can then choose to observe specific patients to see if they are alive or dead. The probability of death is calculated using the following model:

\begin{align}
    \lambda = \exp(\beta \cdot \text{metastasized}) \cdot \lambda_0
    \mu = \text{time\_since\_surgery} \cdot \lambda
\end{align}

The probability of death for a patient is given by the logistic function:

\begin{equation}
p(\text{death}) = \frac{1}{1 + \exp(-\mu)}
\end{equation}

Each patient's outcome is simulated from a Bernoulli distribution with the calculated death probability. The observed data consists of tuples indicating whether the patient died, the time since surgery, and the metastasized status.

For example, for a patient with a given time since surgery and metastasized status, the death outcome is sampled as follows:

\begin{equation}
\text{death\_outcome} \sim \text{Bernoulli}(p(\text{death}))
\end{equation}

\subsection{Predator-Prey Dynamics}

The predator-prey environment models the interaction between populations of predators and prey over time using the Lotka-Volterra equations \citep{volterra1928variations}. The following table describes the inputs and outputs of the experiment:

\begin{table}[h]
\centering
\begin{tabular}{ll}
\toprule
\textbf{Parameter} & \textbf{Description} \\
\midrule
Model & Lotka-Volterra equations \\
Setup Parameters & Initial prey population, initial predator population, $\alpha$, $\beta$, $\gamma$, and $\delta$\\
Observations & Populations of prey and predators at a given time \\
Goals & Predicting populations\\
\bottomrule
\end{tabular}
\vspace{2mm}
\caption{Predator-Prey Environment}
\label{table:predator-prey-experiment}
\end{table}

In this environment, the populations of prey and predators at time $t$ are modeled using the Lotka-Volterra equations. The initial populations of prey and predators are given by the parameters `prey\_init' and `predator\_init', respectively. The interaction between the populations is governed by the parameters $\alpha$, $\beta$, $\gamma$, and $\delta$. The time values range between 0 and 50. The Lotka-Volterra system of differential equations is defined as follows:

\begin{equation}
\frac{d \text{prey}}{dt} = \alpha \cdot \text{prey} - \beta \cdot \text{prey} \cdot \text{predator}
\end{equation}

\begin{equation}
\frac{d \text{predator}}{dt} = \delta \cdot \text{prey} \cdot \text{predator} - \gamma \cdot \text{predator}
\end{equation}

The populations of prey and predators at any given time $t$ are obtained by solving these differential equations. The observed data consists of tuples indicating the time and the populations of prey and predators at that time.

For example, for a given time $t$, the populations of prey and predators are computed by solving the Lotka-Volterra equations with the specified parameters and initial populations. The resulting populations are nonnegative integers representing realistic population counts.

\subsection{Emotions from Outcomes}

The Emotions from Outcomes environment models a participant's predictions of a players emotions after spinning a wheel with three possible monetary outcomes \citep{ong2015affective}. The model considers the actual outcome, the expected outcome, and the absolute difference between the actual and expected outcomes. The following table describes the inputs and outputs of the experiment:

\begin{table}[h]
\centering
\begin{tabular}{ll}
\toprule
\textbf{Parameter} & \textbf{Description} \\
\midrule
Model & Forward regression model with priors for emotional response \\
Setup Parameters & Prize values, probabilities, outcome, LLM \\
Observations & Prediction in natural language of how a player feels and why\\
Goals & Predicting what a participant thinks a player feels on a likert scale of 8 emotions.\\
\bottomrule
\end{tabular}
\vspace{2mm}
\caption{Emotions From Outcomes Environment}
\label{table:emotion-experiment}
\end{table}

In this environment, the participant's predictions of a player's emotions are modelled after observing the outcome of the player spinning a wheel with three possible prizes. Each outcome has a known probability and monetary value. The emotion predictions are influenced by the actual outcome, the difference between the actual outcome and the expected outcome, and the absolute difference between the actual outcome and the expected outcome.

The model uses the following parameters:
\begin{enumerate}
    \item Prize values: The monetary values of the three possible outcomes.
    \item Probabilities: The probabilities of each outcome occurring.
    \item Outcome: The actual outcome of the wheel spin.
\end{enumerate}

The emotions are measured on a Likert scale from 1 to 9 for the following eight emotions:
Happiness,
Sadness,
Anger,
Surprise,
Fear,
Disgust,
Contentment,
Disappointment

The emotional response is generated based on the following model:

\begin{equation}
\text{mean} = \alpha + \beta_{\text{win}} \cdot \text{win} + \beta_{\text{PE}} \cdot \text{PE} + \beta_{\text{absPE}} \cdot \text{absPE}
\end{equation}

where:
\begin{itemize}
    \item $\alpha$ are the intercepts for each emotion.
    \item $\beta_{\text{win}}$ are the coefficients for the actual outcome.
    \item $\beta_{\text{PE}}$ are the coefficients for the prediction error (PE).
    \item $\beta_{\text{absPE}}$ are the coefficients for the absolute prediction error (absPE).
\end{itemize}

For each emotion, the value is sampled from a normal distribution with the computed mean and a predefined standard deviation. 

The generative model produces Likert scale ratings for the 8 emotions for the participant's predictions of what a player would feel. These predictions are translated into free-form natural language observations by a language model with the prompt shown in \autoref{fig:emo-prompt}. For example, an observation when the prizes are \$50, \$20, \$10 with probabilities 0.1, 0.4, 0.5, and the player wins \$50, the simulated participant responds with ``The player might be feeling quite happy and content because they landed on the highest possible outcome, which was unexpected given its low probability.''

\begin{figure*}[ht!]
\centering
\begin{tcolorbox}[
title={LLM prompt to translate predictions from the generative model to observations}]
\fontsize{7pt}{7pt}\selectfont
\ttfamily
\begin{lstlisting}[language={}]
You are observing a user play a game where they spin a wheel.
The wheel has three possible outcomes (monetary values), and the probabilities of landing on each are known to you and the player. 
You are observing the player play the game and the outcomes.
You are asked to predict how the player feels after each spin of the wheel.
Translate the values for emotions to a sentence that describes the player.
The decisions are based on the following model and features:
- Your predition of the player's happiness, sadness, anger, surprise, fear, disgust, contentment, and disappointment are influenced by a few factors.
- The player's emotions are influenced by the actual outcome of the spin.
- The player's emotions are influenced by the difference between the actual outcome and the expected outcome.
- The player's emotions are influenced by the absolute difference between the actual outcome and the expected outcome.
The wheel has three possible outcomes with the following probabilities:
{v1:0.2f}: {p1:0.2f}
{v2:0.2f}: {p2:0.2f}
{v3:0.2f}: {p3:0.2f}
The player has spun the wheel and landed on {outcome}.
This is how you think the player feels:
Happiness: {happiness}/9
Sadness: {sadness}/9
Anger: {anger}/9
Surprise: {surprise}/9
Fear: {fear}/9
Disgust: {disgust}/9
Contentment: {contentment}/9
Disappointment: {disappointment}/9
Translate the values for emotions to a sentence that describes the player.
1: Not at all, 9: Very much
This sentence should be concise and describe the player's emotions after the spin.
The sentence should be a few words long and should not contain any numbers or refer to the numbers directly.
Only talk about the most salient emotions.
Start with "The player might be feeling...because..." and provide a description of the player's emotions and a reason.
\end{lstlisting}
\end{tcolorbox}
\caption{\textbf{LLM prompt for simulated participant.} LLM prompt to translate predictions from the generative model to observations in free-form natural language.
}
\label{fig:emo-prompt}
\end{figure*}
\subsection{Moral Machines}

The Moral Machine environment \citet{awad2018moral} models participants' decisions in moral dilemmas involving autonomous vehicles. Participants are presented with scenarios where the vehicle must decide between two outcomes, each involving the death of a different group of characters. The following table describes the inputs and outputs of the experiment:

\begin{table}[h]
\centering
\begin{tabular}{ll}
\toprule
\textbf{Parameter} & \textbf{Description} \\
\midrule
Model & Logistic regression model with priors for moral decision-making \\
Setup Parameters & Character attributes, intervention type, LLM \\
Observations & Prediction in natural language of which group to save and why \\
Goals & Predicting which group participants choose to save \\
\bottomrule
\end{tabular}
\vspace{2mm}
\caption{Moral Machines Environment}
\label{table:moral-machine-experiment}
\end{table}

In this environment, participants must decide which group of characters to save in a moral dilemma involving autonomous vehicles. The characters in each group can be any of the following: stroller, boy, girl, pregnant\_woman, male\_doctor, female\_doctor, female\_athlete, male\_athlete, female\_executive, male\_executive, large\_woman, large\_man, homeless, old\_man, old\_woman, criminal, dog, cat.

The model uses the following parameters:
\begin{enumerate}
    \item Character attributes: gender, age, social status, fitness, species (human or pet).
    \item Intervention type: 'swerve' or 'stay'.
\end{enumerate}

The decision to save a group is influenced by the difference in attributes between the two groups and the intervention required. The logistic regression model considers the following coefficients:
\begin{itemize}
    \item $\beta_{\text{intervention}}$: Preference for inaction.
    \item $\beta_{\text{group}}$: Preference for group 1 (passengers).
    \item $\beta_{\text{gender}}$: Preference for sparing females.
    \item $\beta_{\text{fitness}}$: Preference for sparing the fit.
    \item $\beta_{\text{social\_status}}$: Preference for sparing higher status individuals.
    \item $\beta_{\text{age}}$: Preference for sparing the young.
    \item $\beta_{\text{human\_count}}$: Preference for sparing more characters.
    \item $\beta_{\text{species}}$: Preference for sparing humans over pets.
\end{itemize}

The logistic regression model generates a choice for which group to save based on the computed attributes and intervention. These predictions are translated into free-form natural language observations by a language model with the prompt shown in \autoref{fig:moral-prompt}. For example, in a scenario where group 1 consists of a boy and a girl, and group 2 consists of an elderly man and an elderly woman, with the intervention being 'swerve', the simulated participant responds with ``I choose to save group 1 because they are younger and have more potential life ahead of them.''

\begin{figure*}[ht!]
\centering
\begin{tcolorbox}[
title={LLM prompt to translate predictions from the logistic regression model to observations}]
\fontsize{7pt}{7pt}\selectfont
\ttfamily
\begin{lstlisting}[language={}]
You have to make choices in a moral dilemma involving autonomous vehicles. In each dilemma, the vehicle must decide between two outcomes, each involving the death of a different group of characters.
There are two groups of characters, and the vehicle must decide which group to save. You can query the participant for their decision by providing the descriptions of the characters in each group and the intervention the car would take ('swerve' or 'stay' to save group 1).
The characters can be any of the following: {characters}
Translate the decision that is shown to you in natural language and provide reasons for it.
The importance that you place on the different attributes of the characters affects your decision. Values range from -1 to 1, with 0 being neutral.
Your preferences are as follows:
- Preference for human count: {beta_human_count}
- Preference for group 1 (passengers): {beta_group}
- Preference for males: {beta_gender}
- Preference for young: {beta_age}
- Social status: {beta_social_status}
- Preference for Fitness: {beta_fitness}
- Preference for humans over pets: {beta_species}
- Preference for having to take an action: {beta_intervention}
DO NOT talk about the values of preferences directly in your response. Only talk about the most salient reasons for the decision. Be concise, write one sentence only.
Translate the decision that is shown to you in natural language and provide reasons for it.
The scenario presented to you was as follows:
Group 1: {group1}
Group 2: {group2}
Intervention: {intervention}
Choice made: {choice}
Translate the decision that is shown to you in natural language and provide reasons for it.
Only talk about the most salient reasons for the decision.
The sentence should be concise and should not refer to the importance factors/values directly.
Start with "I choose to save group 1/group 2 because...". Be concise and faithful to the importance weights.
\end{lstlisting}
\end{tcolorbox}
\caption{\textbf{LLM prompt for simulated participant.} LLM prompt to translate predictions from the logistic regression model to observations in free-form natural language.
}
\label{fig:moral-prompt}
\end{figure*}

\section{Qualitative Examples}
\begin{figure*}[ht!]
\centering
\begin{tcolorbox}[
title={Box's Apprentice proposed programs}]
\fontsize{7pt}{7pt}\selectfont
\ttfamily
\begin{lstlisting}[language={}]
 with pm.Model() as model:
        # Priors for student abilities and question difficulties
        student_ability = pm.Normal("student_ability", mu=0, sigma=1, shape=6)
        question_difficulty = pm.Normal("question_difficulty", mu=0, sigma=1, shape=6)
        
        # Linear combination of student ability and question difficulty
        logit_p = student_ability[student_id] - question_difficulty[question_id]
        
        # Likelihood
        y_obs = pm.Bernoulli("y_obs", logit_p=logit_p, observed=correctness, dims="obs_id")
\end{lstlisting}

\begin{lstlisting}[language=Python]
 with pm.Model() as model:        
        # Define priors for the parameters of the linear model
        intercept = pm.Normal("intercept", mu=0, sigma=10)
        slope = pm.Normal("slope", mu=0, sigma=10)
        
        # Define the linear model for the mean of the Poisson distribution
        lambda_ = pm.math.exp(intercept + slope * year_data)
        
        # Define the likelihood
        y_obs = pm.Poisson("y_obs", mu=lambda_, observed=C, dims="obs_id")
\end{lstlisting}

\begin{lstlisting}[language=Python]
with pm.Model() as model:
        # Define the locations of the three sources as latent variables
        source_x1 = pm.Uniform("source_x1", lower=0, upper=1, shape=3)
        source_x2 = pm.Uniform("source_x2", lower=0, upper=1, shape=3)
        
        # Calculate the distances from each source to each observation point
        distances = pm.math.sqrt((x1_data[:, None] - source_x1[None, :])**2 + 
                                 (x2_data[:, None] - source_x2[None, :])**2)
        
        # Model the signal strength as a function of the distances
        signal_strength_mu = pm.math.sum(1 / distances, axis=1)
        
        # Define the likelihood of the observations
        y_obs = pm.Normal("y_obs", mu=signal_strength_mu, sigma=1, observed=signal_strength, dims="obs_id")
\end{lstlisting}

\end{tcolorbox}
\caption{\textbf{\texttt{BoxLM} proposed programs.} \textbf{(top)} IRT 
\textbf{(middle)} Peregrines 
\textbf{(bottom)} Location finding
}
\label{code:qual}
\end{figure*}

\begin{figure*}[ht!]
\centering
\begin{tcolorbox}[
title={Example of an explanation produced by the LLM Agent for the IRT Environment}]
\fontsize{7pt}{7pt}\selectfont
\ttfamily
\begin{lstlisting}[language={}]
Based on the observations made, we can infer some patterns about the students and questions:

1. **Question 1**: Students 0, 1, 2, and 3 answered question 1 correctly, while student 4 answered it incorrectly. This suggests that question 1 is generally easier for most students, but not universally so.

2. **Question 0**: Students 2, 3, and 4 answered question 0 correctly. This indicates that question 0 might also be relatively easy for students, though we lack data for students 0, 1, and 5.

3. **Student 1**: Answered question 1 correctly but answered question 2 incorrectly. This suggests that student 1's performance may vary depending on the question.

4. **Student 4**: Answered question 0 correctly but answered question 1 incorrectly. This indicates that student 4's performance also varies by question.

From these observations, it appears that some questions (like question 1 and question 0) might be easier for a majority of students, but individual performance can vary. Without more data, it's challenging to make definitive predictions for all student-question pairs, but these patterns provide a starting point for understanding the environment.
\end{lstlisting}
\end{tcolorbox}
\caption{\textbf{Example Explanation.} Example of an explanation produced by the LLM Agent for the IRT Environment.
}
\label{exp:irt2}
\end{figure*}

\begin{figure*}[ht!]
\centering
\begin{tcolorbox}[
title={Example of an explanation produced by Box's Apprentice for the IRT Environment}]
\fontsize{7pt}{7pt}\selectfont
\ttfamily
\begin{lstlisting}[language={}]
Based on the observations and the statistical model, we can infer that each student's ability and each question's difficulty play significant roles in determining the correctness of answers. The model uses these factors to predict the probability of a correct response.

**Student Abilities:**
- Students have varying abilities, with some being more likely to answer correctly than others. For instance, student 2 has a higher ability (0.544) compared to student 0 (-0.405).

**Question Difficulties:**
- Questions also vary in difficulty. For example, question 1 is relatively easier (-0.873) compared to question 3 (0.492).

**Predicted Probabilities:**
- The model calculates the probability of a correct answer by combining student ability and question difficulty. For example, student 2 answering question 1 has a high probability of correctness (0.708), while student 0 answering question 3 has a lower probability (0.414).

**General Patterns:**
- Students with higher abilities are more likely to answer correctly across various questions.
- Easier questions are more likely to be answered correctly by most students.

To predict if a student will answer a question correctly, consider both the student's ability and the question's difficulty. Higher student ability and lower question difficulty increase the likelihood of a correct answer.
\end{lstlisting}
\end{tcolorbox}
\caption{\textbf{Example Explanation.} Example of an explanation produced by the Box's Apprentice for the IRT Environment.
}
\label{exp:irt}
\end{figure*}

\end{document}